\documentclass[11pt]{article}

\usepackage[final]{acl}
\usepackage{times}
\usepackage{latexsym}
\usepackage[T1]{fontenc}
\usepackage[utf8]{inputenc}
\usepackage{microtype}
\usepackage{inconsolata}
\usepackage{graphicx}
\usepackage{amsmath}
\usepackage{booktabs}
\usepackage{enumitem}

\newif{\ifhidecomments}
\ifhidecomments
    \newcommand{\viv}[1]{}
    \newcommand{\chenhao}[1]{}
    \newcommand{\zana}[1]{}
\else
    \newcommand{\chenhao}[1]{\textcolor{red}{[#1 ---\textsc{chenhao}]}}
    \newcommand{\viv}[1]{\textcolor{blue}{[#1 ---\textsc{viv}]}}
    \newcommand{\zana}[1]{\textcolor{orange}{[#1 ---\textsc{zana}]}}
\fi

\title{Users Mispredict Their Own Preferences for AI Writing Assistance}

\author{
    \textbf{Vivian Lai}\textsuperscript{1}\quad
    \textbf{Zana Bu\c{c}inca}\textsuperscript{2}\\
    \textbf{Nil-Jana Akpinar}\textsuperscript{1}\quad
    \textbf{Mo Houtti}\textsuperscript{1}\quad
    \textbf{Hyeonsu B. Kang}\textsuperscript{1}\\
    \textbf{Kevin Chian}\textsuperscript{1}\quad
    \textbf{Namjoon Suh}\textsuperscript{1}\quad
    \textbf{Alex C. Williams}\textsuperscript{1}\\[4pt]
    \textsuperscript{1}\,Microsoft \qquad
    \textsuperscript{2}\,Massachusetts Institute of Technology\\[2pt]
    \texttt{viv.lai@microsoft.com}
}

\begin{document}
\maketitle
\begin{abstract}
Proactive AI writing assistants need to predict when users want drafting help, yet we lack empirical understanding of what drives preferences. Through a factorial vignette study with 50 participants making 750 pairwise comparisons, we find compositional effort dominates decisions ($\rho = 0.597$) while urgency shows no predictive power ($\rho \approx 0$). More critically, users exhibit a striking perception-behavior gap: they rank urgency first in self-reports despite it being the weakest behavioral driver, representing a complete preference inversion. This misalignment has measurable consequences. Systems designed from users' stated preferences achieve only 57.7\% accuracy, underperforming even naive baselines, while systems using behavioral patterns reach significantly higher 61.3\% ($p < 0.05$). These findings demonstrate that relying on user introspection for system design actively misleads optimization, with direct implications for proactive natural language generation (NLG) systems.
\end{abstract}

\section{Introduction}
\label{sec:intro}

LLMs have enabled increasingly capable AI writing assistants, yet a fundamental design question remains unresolved: When should systems proactively offer drafting help? While reactive assistance responds to explicit user requests, proactive systems need to infer opportune moments from context. Prior work indicates that well‑timed initiative, grounded in the costs and benefits of action under uncertainty, can improve user experience, whereas poor timing raises interruption costs and user frustration~\cite{horvitz1999principles, iqbal2008effects, mcfarlane2002comparison}. Despite widespread deployment of such systems in email clients (e.g., Smart Reply~\cite{kannan2016smart}, Smart Compose~\cite{chen2019gmail}), designers still lack robust evidence on which contextual cues reliably predict actual desire for assistance. Moreover, stated preferences and subjective measures often misalign with observed behavior and performance, and suggestion UIs can shift composition, introducing trade-offs for proactive assistance~\cite{bucinca2020proxy,poursabzi2021manipulating,buschek2021impact}.

A common intuition is to offer assistance under time pressure, but interruption research emphasizes task structure and breakpoints over urgency as better predictors of lower disruption timing~\cite{adamczyk2004if,iqbal2008effects}. An alternative is to expose user controls and preference settings, consistent with human‑AI interaction guidelines~\cite{amershi2019guidelines}, while recognizing that self‑reports may not faithfully reflect assistance needs~\cite{bucinca2020proxy}.

We investigate these questions through email drafting, a domain that offers several methodological advantages for studying proactive assistance preferences. Email represents a high-frequency writing task with measurable contextual factors and clear ecological validity as a deployed AI assistance application. Email drafting exemplifies the core challenge facing proactive NLG systems, namely inferring when users want assistance from contextual signals alone without explicit requests.

Through a factorial vignette study, we examine what email attributes predict AI assistance preferences. We manipulate four dimensions: urgency, compositional effort, sender importance, and email type, across 16 scenarios. Fifty participants made 750 pairwise comparisons and reported their stated preference drivers. We address three research questions: (1) What email attributes drive preferences for AI drafting assistance? (\S\ref{sec:rq1}) (2) Do users accurately perceive what drives their preferences for AI assistance? (\S\ref{sec:rq2}) (3) Do ML models predict preferences from behavioral patterns, and what is the cost of the perception-behavior gap? (\S\ref{sec:rq3})

Our findings reveal a striking misalignment. Users believe urgency is most important (mean self-reported rank: 2.06 out of 4), yet behavioral analysis shows urgency has near-zero predictive power ($\rho \approx 0$). Instead, compositional effort dominates ($\rho$ = 0.597). This represents a complete preference inversion that reveals poor metacognitive accuracy: urgency ranks first in stated preferences but last behaviorally, while effort shows the opposite pattern, demonstrating users cannot accurately perceive what drives their own preferences.

To quantify the misalignment, we train ML models to predict individual preferences and compare three rule-based strategies, namely uniform weighting, stated-preference weighting from user surveys, and behavioral weighting from revealed preferences. The stated-preference approach achieves only 57.7\% accuracy, worse than the naive baselines, while behavioral weighting reaches significantly higher 61.3\% ($p < 0.05$).


We make the following three contributions:

\begin{enumerate}[itemsep=5pt, parsep=0pt, topsep=0pt]
\item We provide empirical evidence that compositional effort, not urgency, drives AI drafting assistance preferences. This reorients design priorities from temporal signals toward cognitive load detection.

\item We show a perception-behavior gap where users misidentify their preferences, revealing poor metacognitive accuracy. This demonstrates that NLG systems should learn from implicit behavioral feedback rather than explicit user surveys.

\item We quantify the cost of the perception-behavior gap, showing that behavioral patterns achieve 61.3\% prediction accuracy, significantly outperforming stated-preference designs at 57.7\% (p < 0.05). This demonstrates the practical cost of designing from user self-reports rather than observed behavior.
\end{enumerate}

This work advances human-centered AI by grounding proactive assistance in behavior, revealing limits of user introspection, and validating insights for deployment.
\section{Related Work}
\label{sec:related_work}

\noindent\textbf{Email management \& proactive AI assistance.} Research on email management has long argued for and designed more intelligent assistance~\cite{whittaker1996email,bellotti2003taking}, establishing that users face systematic challenges in handling email volume and allocating response effort~\cite{dabbish2005email}. Synthesizing prior work, we focus on four contextual factors that influence email behavior: urgency~\cite{tyler2003expect}, compositional effort~\cite{bellotti2003taking,whittaker1996email}, sender importance~\cite{dabbish2005email}, and email type~\cite{mackay1988diversity,whittaker1996email,ducheneaut2001mail}. These dimensions inform our factorial design while representing practically relevant features for AI assistance systems.

Recent advances in LLMs have enabled proactive writing assistance~\cite{lee2022coauthor,buschek2021impact}. When timed to natural breakpoints, proactive assistance reduces reaction time and frustration and lowers interruption costs~\cite{horvitz1999principles,cutrell2001notification,iqbal2008effects}. However, poorly timed assistance disrupts workflow~\cite{iqbal2010notifications,czerwinski2004diary,cutrell2000effects}. Although many deployed tools remain reactive, even prominent email assistants~\cite{chen2019gmail,kannan2016smart} rarely report empirically validated timing policies. Our work provides the first systematic investigation of what contextual factors drive users' desire for AI drafting assistance.

\noindent\textbf{User perception in AI systems.} Recent research highlights mismatches between what users believe and how they actually behave with AI systems. 
Evaluations based on proxy tasks and subjective measures fail to predict real decision performance~\cite{bucinca2020proxy}. Human-AI interaction is especially challenging due to uncertainty about AI capabilities and complex outputs~\cite{yang2020reexamining}. While established design guidelines emphasize user control and opportunities to understand or correct system behavior~\cite{amershi2019guidelines}, such approaches can fall short when users misconstrue their needs or an AI’s capabilities. In AI writing assistance, presenting multiple suggestions changes composition behavior and perceived usefulness across user groups~\cite{buschek2021impact}. However, prior work has not manipulated contextual factors nor quantified system consequences. We address these gaps through factorial methodology and computational validation.
\section{Methods}
\label{sec:methods}

\noindent\textbf{Study design.} We employed a $2 \times 2 \times 2 \times 2$ factorial vignette design manipulating four binary email dimensions: urgency, compositional effort, sender importance, and email type, yielding 16 scenarios generated via GPT-5 for consistency (Appendix~\ref{subsec:scenarios}). Fifty participants who regularly handle email in professional contexts completed 15 pairwise comparisons each, randomly sampled from all 120 possible pairs ($\binom{16}{2}$), selecting which scenario they preferred AI drafting assistance for and providing justifications. An exit survey collected rankings of the four dimensions, AI usage patterns, and demographics. See Appendix~\ref{subsec:exit_survey} for study interface and complete exit survey.

\noindent\textbf{Analytical approach.} We used Bradley-Terry modeling to derive preference strengths from pairwise comparisons, then calculated Spearman correlations to identify behavioral drivers (RQ1). We validated findings with ordinal logistic regression and two-way ANOVAs. For RQ2, we computed metacognitive accuracy as the correlation between each participant's stated and revealed preferences, testing predictors via ANOVAs and t-tests. To complement quantitative analysis, we conducted LLM-assisted thematic coding of 750 justifications and 50 survey reflections (Appendix~\ref{subsec:thematic_methods}).

For RQ3, we trained four model classes (Logistic Regression, Random Forest, Gradient Boosting, Neural Network) using 5-fold cross-validation with shallow architectures to prevent overfitting. We compared three fixed-weight strategies (Uniform, Stated Preference, Behavioral) to quantify the perception-behavior gap cost. Complete specifications appear in Appendix~\ref{subsec:ml_details}.
\section{What email attributes drive users' preferences for AI drafting assistance?}
\label{sec:rq1}

\begin{figure}[t]
\centering
\includegraphics[width=0.48\textwidth]{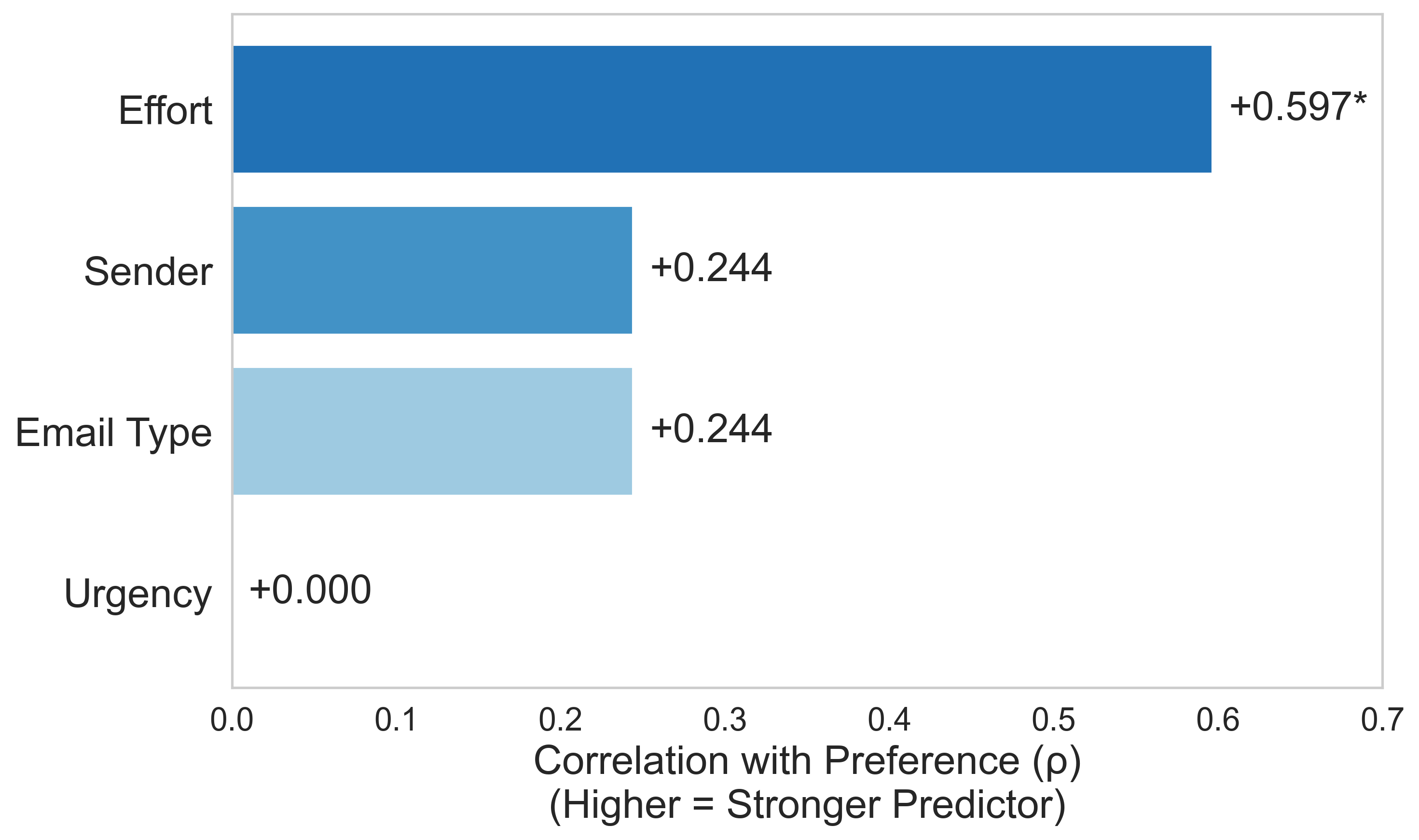}
\vspace{-1.5em}
\caption{Predictive strength of email dimensions measured by Spearman correlation with Bradley-Terry preference strengths. Effort to compose dominates as the only statistically significant predictor, while urgency shows near-zero correlation. Sender importance and email type show weak non-significant trends. *$p<0.05$.}
\vspace{-1.5em}
\label{fig:dimension_effects}
\end{figure}

\begin{figure}[t]
\centering
\includegraphics[width=0.48\textwidth]{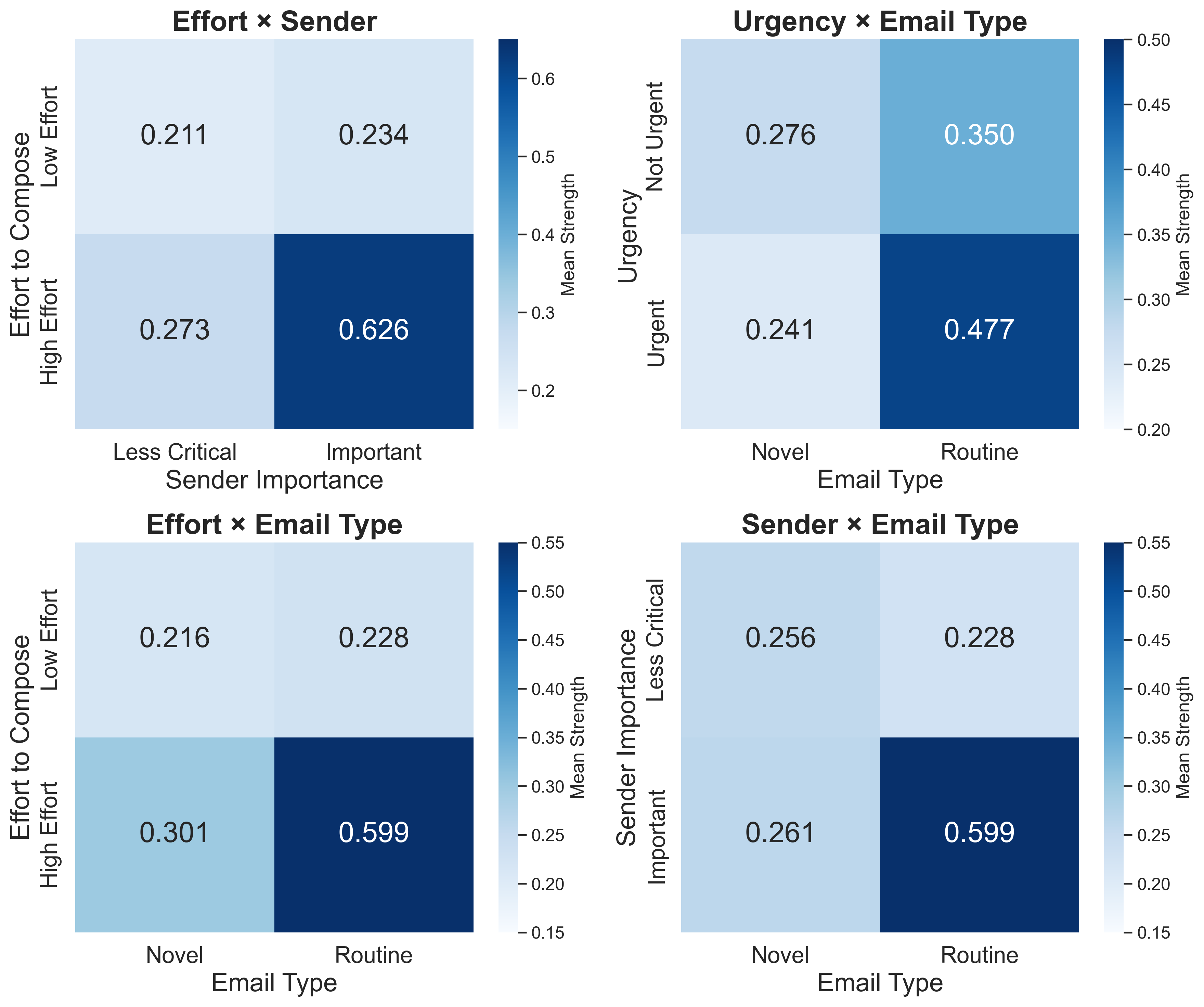}
\vspace{-1.5em}
\caption{Key pairwise dimension interactions showing mean Bradley-Terry preference strengths. Effort $\times$ Sender produces the highest preference when both factors align (important sender $\times$ high effort: $\bar{\pi}$ = 0.626), demonstrating a ``high-stakes, high-burden" heuristic. Sender $\times$ Type shows sender importance primarily matters for routine emails ($\bar{\pi}$ = 0.599) with minimal effect on novel emails.}
\vspace{-1.5em}
\label{fig:dimension_interactions}
\end{figure}

We derive Bradley-Terry preference strengths for all 16 scenarios from 750 pairwise comparisons by 50 participants. Spearman correlations between dimensions and these scenario-level strengths reveal that effort dominates AI drafting preferences ($\rho = 0.597, p < 0.05$), while urgency shows virtually no predictive power ($\rho \approx 0, p = 1.0$). Sender importance ($\rho = 0.244$, favoring important sender) and email type ($\rho = 0.244$, favoring routine over novel email) show weak non-significant trends and play secondary roles (Figure~\ref{fig:dimension_effects}). Ordinal regression validates these patterns. High-effort emails show 13.68 times higher odds of selection ($p < 0.05$), while urgency exhibits no significant effect. Both analyses confirm that effort is the main driver and urgency has no reliable effect.

These dimensions do not operate in isolation and exhibit strong interaction effects (Figure~\ref{fig:dimension_interactions}). High effort combined with important sender produces the highest preference ($\bar{\pi} = 0.626$), suggesting users employ a multiplicative ``high-stakes, high-burden'' heuristic. Conversely, urgency shows context-dependent effects. Users prefer AI help for urgent routine emails ($\bar{\pi} = 0.477$) but not urgent novel emails ($\bar{\pi} = 0.241$), perhaps reflecting greater trust in AI for predictable responses. While 2-way ANOVA confirms effort's main effect ($F = 5.294, p < 0.05$), the interaction effects do not reach statistical significance. However, ML feature importance analysis (\S\ref{subsec:predictive}), which uses all 750 individual comparisons rather than 16 aggregated scenario means, reveals these interactions account for 61.7\% of predictive variance. This suggests genuine interaction structure that traditional ANOVA lacks statistical power to detect.

\begin{figure*}[t]
\centering
\includegraphics[width=0.8\textwidth]{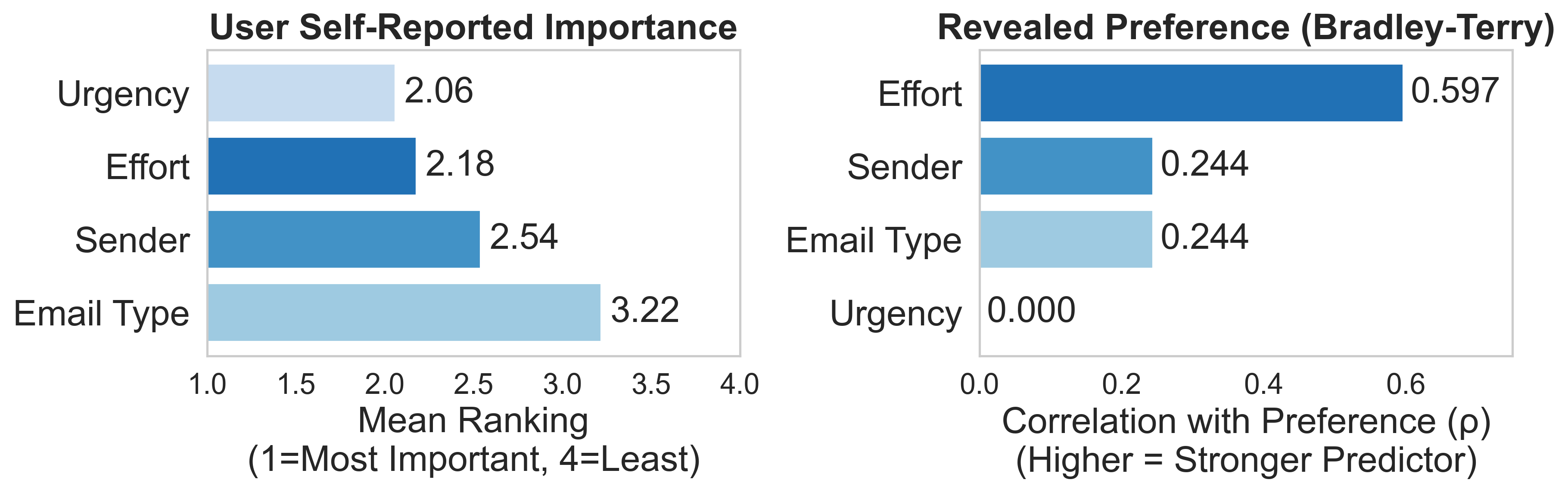}
\vspace{-1em}
\caption{Comparison of stated versus revealed preference rankings. Users report urgency as most important (rank 2.06) but behavioral analysis shows it has no predictive power ($\rho \approx 0$). Effort, the actual strongest driver ($\rho = 0.597$), receives only moderate stated importance (rank 2.18). The non-significant correlation masks a systematic inversion at the extremes, suggesting stated preferences will prioritize the wrong factors.}
\vspace{-1em}
\label{fig:ranking_comparison}
\end{figure*}


\subsubsection{Qualitative Analysis of Justifications}
\label{subsubsec:qualitative_justifications}

Thematic analysis of 750 in-the-moment justifications (avg. 1.03 themes per response) both corroborates and complicates the quantitative patterns. \textbf{Urgency and time pressure} was the most frequently mentioned factor (20.3\%), with users citing deadlines and time pressure as triggers for AI assistance. This creates a puzzle that users articulate urgency most often yet it shows near-zero behavioral predictive power. \textbf{Complexity of drafting} was the second-most frequent theme (14.7\%), with users explicitly seeking AI for multi-step, synthesis-heavy tasks and building comprehensive transition plans, aligning with effort's dominant behavioral correlation ($\rho = 0.597$). Users also actively rejected AI for simple yes/no responses (\textbf{Task simplicity – doesn't need AI}, 12.8\%), confirming that compositional burden drives preferences.

Co-occurrence analysis reveals sophisticated decision heuristics that explain these patterns. Users calibrate AI trust based on task predictability: \textbf{Template or historical context} (10.1\%) frequently co-occurred with perceived AI capability, with users stating ``format already existing so human validation will be faster.'' This conditional trust pattern suggests templates simultaneously reduce effort \textit{and} increase confidence in AI accuracy. Some users also mentioned \textbf{Ease of verification} (1.3\%) when justifying AI use, with reasoning like ``It's easy for me to verify correctness.'' Here, verification refers to checking the AI-generated draft before sending. Users are more willing to adopt AI when they can quickly review outputs to catch any errors. This reframes the design challenge from maximizing AI accuracy to minimizing verification friction, suggesting features like confidence scores or highlighting AI-generated content could expand adoption among risk-averse users.

One surprising insight is that \textbf{Avoiding high-risk or sensitive tasks} was rare (4.8\%), suggesting sensitivity concerns do not dominate AI adoption decisions when verification is feasible. Additionally, users view AI as a cognitive partner for \textbf{Brainstorming or creative help} (3.9\%), not just automation, indicating potential for collaborative features beyond drafting. See Appendix~\ref{subsec:thematic_findings} for theme frequencies and additional insights.
\section{Do users accurately perceive what drives their preferences for AI assistance?}
\label{sec:rq2}

Comparing self-reported rankings with revealed preferences reveals a perception-behavior gap (Figure~\ref{fig:ranking_comparison}). Users rank urgency as most important (mean rank: 2.06) despite its near-zero behavioral predictive power ($\rho \approx 0$). Effort, the strongest driver ($\rho = 0.597$), receives only moderate self-reported importance (mean rank: 2.18). This represents a preference inversion at the extremes, where users' top-ranked dimension has no behavioral effect while the actual strongest driver receives middling stated importance. This shows that designing from stated preferences would prioritize the weakest driver while underweighting the strongest.

\subsection{Qualitative Analysis of Post-Task Reflections}

Thematic analysis of 50 survey responses (avg. 2.8 themes per response) corroborates this perception-behavior gap. \textbf{Urgency / time sensitivity} dominated retrospective accounts at 66\% of responses, representing massive over-reporting compared to immediate justifications at 20.3\% and behavioral reality ($\rho \approx 0.000$). In contrast, \textbf{Effort vs. efficiency} appeared in only 32\% of reflections, understated relative to its actual dominance ($\rho = 0.597$). This pattern of over-reporting urgency and under-reporting effort in retrospective accounts contrasts sharply with behavioral reality. 

Beyond the urgency-effort divergence, the survey reflections reveal how users conceptualize their decision-making. \textbf{Data availability} emerged in 28\% of accounts despite not being a manipulated dimension, reflecting how users decompose compositional effort into information-gathering versus drafting complexity. The key evidence for misperception remains the systematic over-reporting of urgency and under-reporting of effort. See Appendix~\ref{subsec:thematic_findings} for complete thematic comparison.

\subsection{Individual Differences in Metacognitive Accuracy}

We explore whether certain user characteristics show better alignment between stated and revealed preferences. Individual differences reveal that self-reported comfort with AI technology (measured on a 7-point scale), rather than usage frequency or email volume, predicts metacognitive accuracy. High-comfort users (scores 5-7) show better alignment between stated and actual preferences ($\rho = 0.090$), while low-comfort users (scores 1-3) show inverted understanding ($\rho = -0.550$, $p < 0.05$). This misperception manifests in dimension rankings: low-comfort users rank urgency as their top priority (mean rank: 1.00) while placing effort last (3.25), creating a 2.25-rank gap. In contrast, high-comfort users show nearly balanced rankings (urgency: 2.08, effort: 2.05), suggesting accurate understanding of effort's importance. 

Neither AI usage frequency nor email volume significantly predicts metacognitive accuracy, indicating psychological readiness matters more than behavioral experience (see Figure~\ref{fig:metacognitive_accuracy} in Appendix~\ref{subsec:appendix_individual_differences}). That is, users comfortable with AI technology show better self-awareness of their preference drivers, while discomfort amplifies the urgency-over-effort misperception regardless of how frequently they use AI or handle emails. 
\section{Do ML models predict preferences from behavioral patterns, and what is the cost of the perception-behavior gap?}
\label{sec:rq3}

The behavioral findings from \S\ref{sec:rq1} and \S\ref{sec:rq2} establish that users prefer AI drafting assistance for high-effort emails yet believe urgency drives their decisions, revealing a substantial perception-behavior gap. For these insights to inform practical NLG systems, two critical questions remain. First, can ML models accurately predict preferences from email attributes? Second, what is the cost of the perception-behavior gap? How do systems designed from stated beliefs compare with behaviorally-grounded alternatives? We address these questions through ML validation (\S\ref{subsec:predictive}) and system design comparison (\S\ref{subsec:design_comparison}).

\begin{figure}[t]
\centering
\includegraphics[width=0.48\textwidth]{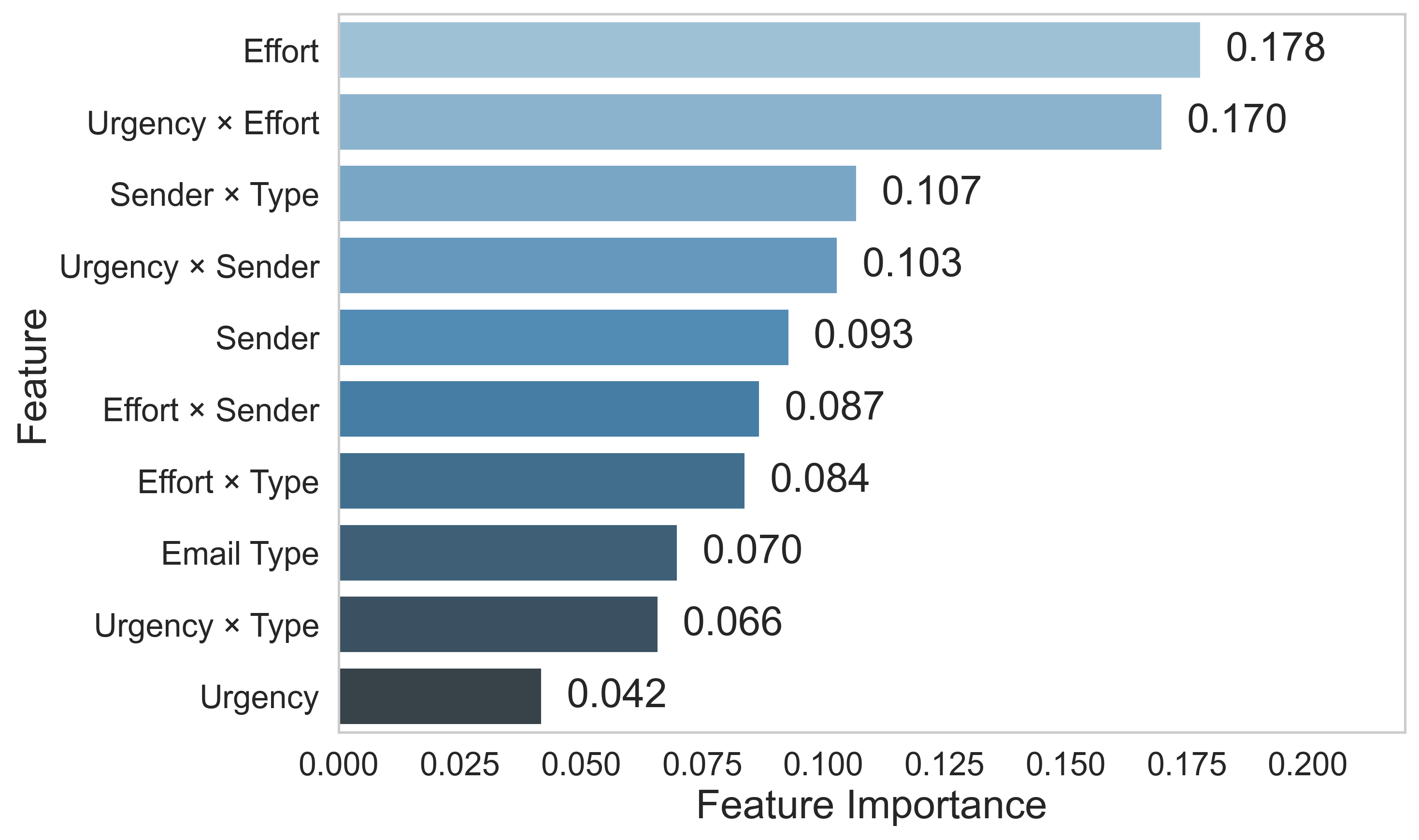}
\caption{Feature importance for preference prediction using Random Forest with all pairwise interactions. Effort main effect (17.8\%) is the single most important feature, followed closely by Effort $\times$ Urgency interaction (17.0\%). Interactions collectively dominate at 61.7\% versus 38.3\% for main effects, demonstrating synergistic preference structure. However, models trained with interactions (59.5\% accuracy) perform worse than models using only main effects (59.7\%), indicating interactions contribute to feature importance without improving predictive performance.}
\label{fig:feature_importance}
\end{figure}

\subsection{Predictive Validation}
\label{subsec:predictive}
We first established baseline predictive performance using only the 4 main effect features. All models achieved approximately 60\% prediction accuracy (Table~\ref{tab:ml_performance}), exceeding the majority class baseline (55.2\%) and significantly outperforming random chance ($p < 0.001$). To test whether interactions improve prediction beyond this baseline, we then trained models with all 10 features (4 main effects + 6 pairwise interactions). Surprisingly, adding interactions provides no predictive benefit: models trained with all 10 features achieve 59.5\% mean accuracy versus 59.7\% for main effects alone.

\begin{table}[t]
\centering
\small
\begin{tabular}{l@{\hskip 0.3cm}c@{\hskip 0.25cm}c@{\hskip 0.5cm}c@{\hskip 0.25cm}c}
\toprule
& \multicolumn{2}{c}{\textbf{Full Features (10)}} & \multicolumn{2}{c}{\textbf{Main Effects (4)}} \\
\cmidrule(r){2-3} \cmidrule(l){4-5}
Model & Acc. & F1 & Acc. & F1 \\
\midrule
Log. Reg. & 0.587 & 0.657 & 0.580 & 0.643 \\
Random Forest & 0.608 & 0.719 & 0.605 & 0.714 \\
Grad. Boost & 0.588 & 0.709 & 0.600 & 0.713 \\
Neural Net & 0.599 & 0.692 & 0.603 & 0.698 \\
\midrule
\textbf{Mean} & \textbf{0.595} & \textbf{0.694} & \textbf{0.597} & \textbf{0.692} \\
\bottomrule
\end{tabular}
\caption{ML model performance comparison between full feature set (4 main effects + 6 pairwise interactions) and main effects only. All models trained with 5-fold cross-validation. Main effects achieve slightly higher mean accuracy (59.7\% vs 59.5\%), demonstrating interactions provide no predictive benefit. Full metrics available in Appendix~\ref{subsec:ml_details}.}
\label{tab:ml_performance}
\end{table}

This result appears paradoxical given that feature importance analysis reveals interactions dominate predictive importance. When we examine the Random Forest model (the best-performing model) trained with all 10 features, interactions contribute 61.7\% of feature importance versus 38.3\% for main effects (Figure~\ref{fig:feature_importance}). Effort's dominance manifests in both its main effect (17.8\%, the single most important feature) and its interactions, particularly Effort $\times$ Urgency (17.0\%), Effort $\times$ Sender (8.7\%), and Effort $\times$ Type (8.4\%). Beyond effort, other notable interactions include Sender $\times$ Type (10.7\%) and Urgency $\times$ Sender (10.3\%).

How can interactions dominate feature importance yet provide no predictive benefit? The resolution lies in recognizing that interactions capture variance \textit{already explained} by main effects in our binary feature space. Since effort appears in multiple high-importance interactions (Effort $\times$ Urgency, Effort $\times$ Sender, Effort $\times$ Type), its predictive signal gets counted multiple times across different interaction terms. When effort is high, many of its interactions will also be non-zero, but this redundancy provides no new predictive information beyond the main effect itself. The modest performance difference suggests that in our experimental setting with binary features, main effects alone efficiently capture the preference structure.

\begin{table}[t]
\centering
\small
\begin{tabular}{lcccc}
\toprule
Strategy & Acc. & Prec. & Recall & F1 \\
\midrule
Uniform & 0.589 & 0.586 & 0.870 & 0.700 \\
Stated Preference & 0.577 & 0.568 & 0.981 & 0.719 \\
Best ML Model & 0.605 & 0.614 & 0.855 & 0.714 \\
\textbf{Behavioral} & \textbf{0.613} & \textbf{0.608} & \textbf{0.841} & \textbf{0.706} \\
\bottomrule
\end{tabular}
\caption{Performance comparison of rule-based strategies and best ML model (Random Forest with 4 main effects). Simple behavioral rules achieve parity with ML models, demonstrating that correct feature weighting matters more than algorithmic sophistication. Both behavioral and ML significantly outperform stated preference in accuracy (behavioral \& ML: $p < 0.05$); F1 differences are not statistically significant. McNemar's tests with continuity correction.}
\label{tab:strategy_comparison}
\end{table}

\subsection{System Design Strategy Comparison}
\label{subsec:design_comparison}
To evaluate the cost of the perception-behavior gap, we compared three rule-based weighting strategies against the best ML model. Since main effects alone are sufficient and interactions provide no predictive benefit, we use Random Forest with 4 main effects (60.5\% accuracy, comparable to the 10-feature version at 60.8\%), as shown in Table~\ref{tab:strategy_comparison}. Table~\ref{tab:strategy_weights} shows the feature weights for each strategy.

\begin{table}[t]
\centering
\small
\begin{tabular}{lcccc}
\toprule
Strategy & Urgency & Effort & Sender & Type \\
\midrule
Uniform & 25\% & 25\% & 25\% & 25\% \\
Stated Preference & 40\% & 30\% & 20\% & 10\% \\
Behavioral & 0\% & 55\% & 22.5\% & 22.5\% \\
\bottomrule
\end{tabular}
\caption{Feature weights for rule-based strategies. Uniform assigns equal importance. Stated Preference reflects user survey rankings. Behavioral reflects Bradley-Terry correlation strengths, with urgency near-zero.}
\label{tab:strategy_weights}
\end{table}

The stated-preference strategy achieved 57.7\% accuracy, underperforming baseline at 58.9\%, demonstrating the perception-behavior gap from \S\ref{sec:rq2}. In contrast, behavioral weighting achieved 61.3\% accuracy, a 3.6-point improvement over stated preference ($p < 0.05$). The best ML model achieved 60.5\% accuracy, significantly better than stated preference ($p < 0.05$) and uniform baseline ($p < 0.05$), but statistically indistinguishable from behavioral weighting. 

Two key insights emerge from this comparison. First, the perception-behavior gap has measurable costs. Stated preference's poor performance stems from inverted weights: allocating 40\% to urgency despite its zero predictive power, while behavioral weighting correctly prioritizes effort at 55\%. This weight inversion causes stated preference to significantly underperform both behavioral and ML approaches. Second, simple behavioral rules achieve parity with ML models, demonstrating that correct feature weighting matters more than algorithmic sophistication. Once weights are correct, ML provides no additional benefit, with both behavioral rules and ML achieving approximately 61\% accuracy (performance ceiling discussed in Limitations, \S\ref{sec:limitations}). These findings have implications for deployment strategy, discussed in \S\ref{subsec:design_implications}.
\section{Discussion}
\label{sec:discussion}

\subsection{Why Do Users Misperceive Their Own Preferences?}

The perception-behavior gap documented in \S\ref{sec:rq2} raises a fundamental question: Why do users misidentify urgency as their primary decision factor when it has no predictive power? We propose two complementary explanations. 

\noindent\textbf{Social desirability bias}~\cite{paulhus1991measurement,crowne1960new} \textbf{may drive explicit rankings toward normatively acceptable justifications.} Admitting ``I want AI help to avoid cognitive effort'' sounds less professional than ``I need help with time-sensitive emails.'' Related work in HCI and survey methodology shows that subjective measures and stated preferences can misalign with behavior and mislead evaluation~\cite{bucinca2020proxy,bertrand2001people}. Supporting this interpretation, thematic analysis of post-task survey reflections revealed that users frequently mentioned quality and personalization concerns even though these factors did not predict which emails they actually preferred AI assistance for. This pattern suggests users invoke professionally acceptable justifications to maintain their identity as careful, quality-conscious professionals, even when these concerns do not actually drive their assistance preferences. 

\noindent The \textbf{availability heuristic}~\cite{tversky1973availability} means that more salient or emotionally charged events, such as urgent scenarios, are easier to recall and may be overemphasized in people's judgments. Participants may recall urgent emails because time pressure creates stress, making these episodes cognitively salient even if urgency had no actual influence on their preference decisions. In contrast, high-effort emails may feel routine despite requiring more support, leading users to underestimate effort's influence. The retrospective elaboration pattern documented in \S\ref{sec:rq2}, where urgency mentions rose from 20.3\% to 66\% while effort mentions remained understated, provides empirical support for this reconstructive process.

These two mechanisms likely operate in sequence and reinforce each other. When asked to retrospectively explain their preferences, participants first retrieve accessible memories through the availability heuristic. Urgent emails are emotionally charged and create stress, making them cognitively salient and easily recalled. In contrast, high-effort emails may feel routine despite requiring substantial cognitive resources, causing participants to underestimate effort's influence. Once urgency is retrieved as salient, social desirability bias then shapes how participants report these memories. Framing assistance needs around urgency (``helping me meet deadlines'') sounds more professionally acceptable than admitting effort-avoidance (``helping me avoid cognitively demanding work''), particularly for knowledge workers whose professional identity centers on handling complex tasks.

The individual differences in metacognitive accuracy documented in \S\ref{sec:rq2} support this dual-mechanism interpretation. AI comfort moderates the gap not by changing actual preferences, but by reducing social desirability pressure. Users comfortable with AI technology feel less need to justify assistance-seeking through normatively acceptable urgency framing. Conversely, low-comfort users may experience amplified effects from both mechanisms: heightened anxiety increasing urgency's memorability while uncertainty about AI appropriateness intensifying impression management concerns. This explains why behavioral experience (usage frequency, email volume) shows no relationship with metacognitive accuracy, while psychological readiness does. Comfort with AI technology reduces the distorting effects of both availability-based retrieval and socially filtered reporting.

\subsection{Design Implications for Proactive NLG Systems}
\label{subsec:design_implications}

Our findings challenge a foundational assumption in human-centered NLG: explicit user feedback reliably guides system optimization. Stated-preference weighting underperformed naive baselines, demonstrating that designing from what users say yields objectively worse predictions than behavioral approaches. We propose three core principles for proactive NLG system design.

\noindent\textbf{Cognitive load detection over temporal urgency.} Compositional effort dominates preference decisions while urgency shows no predictive power, contradicting common design intuitions to offer assistance under time pressure. Systems should detect signals of compositional complexity rather than temporal signals. Qualitative analysis reveals users seek assistance for multi-step, synthesis-heavy tasks (14.7\% of justifications) while actively rejecting AI for simple responses (12.8\%). This suggests proactive systems should identify task structure indicators such as response complexity, coordination requirements across multiple stakeholders, and synthesis needs rather than deadline proximity or email flagging as urgent.

While our study establishes that effort drives preferences, operationalizing effort detection in production systems remains an open challenge. Potential signals include thread length and quoted content to indicate synthesis requirements, recipient diversity to detect coordination complexity, and question structure to distinguish simple confirmations from multi-part responses. The interaction between effort and sender importance suggests systems should apply contextual weighting rather than uniform thresholds, prioritizing effort detection more heavily for high-stakes communication. Future work should validate which specific signals best predict compositional burden across different communication contexts.

\noindent\textbf{Implicit behavioral feedback over explicit preference elicitation.} Stated preferences actively mislead optimization. Settings pages asking users to specify when they want AI assistance assume introspective access to decision drivers, which our findings show is unreliable. Systems should learn from implicit usage patterns such as which suggestions users accept, how long they deliberate before responding, and how frequently they override recommendations rather than upfront configuration.

Beyond learning from usage signals, systems should proactively surface contextual cues that reduce adoption friction. When offering assistance, highlight template availability or similar past emails to simultaneously signal lower effort requirements and increase confidence in AI output quality. This conditional trust pattern (\S\ref{subsubsec:qualitative_justifications}) suggests that assistance prompts framed as "I can draft this using your previous response to [similar email]" may be more effective than generic "Would you like help?" offers.

However, implicit learning without transparency risks manipulating user trust. Showing users their revealed preferences serves dual purposes: providing accountability for system decisions while helping users understand when they genuinely need assistance. This transparency is particularly valuable because users consistently misidentify their preference drivers (\S\ref{sec:rq2}). Low-comfort users, who exhibit the most extreme misperception between stated and revealed preferences, may benefit most from seeing their actual behavioral patterns.

\noindent\textbf{Simple behavioral rules before complex modeling.} The equivalence between behavioral rules and ML has implications for system deployment. For initial launches, behavioral rules offer three advantages over learned models. First, they require no training data, enabling deployment before usage logs accumulate. This is critical for cold-start scenarios where systems must make predictions immediately. Second, they provide interpretability and debuggability. Stakeholders can audit the rule ``weight effort 55\%, sender/type 22.5\% each, urgency near zero'' and understand why the system behaves, facilitating trust and allowing manual refinement. Third, they avoid overfitting in sparse-data regimes. Behavioral weights derived from aggregate analysis across users encode population-level preference structure, whereas learned ML weights can overfit to individual idiosyncrasies when per-user data is limited.

However, as systems mature and accumulate usage data, ML may offer advantages. Personalization becomes feasible when individual users generate sufficient implicit feedback signals. Context-specific adaptation can capture how preferences vary by time of day, workload, or project phase. The key insight is sequencing: establish correct baseline weights from behavioral analysis first, then layer ML personalization incrementally. Deploying ML prematurely, or worse, encoding inverted stated-preference weights, locks in poor performance that future usage data struggles to correct. The 2.8 to 3.6 gap between stated preference and both behavioral/ML approaches demonstrates that wrong initial weights degrade performance.
\section{Conclusion}
\label{sec:conclusion}

Proactive NLG systems need to infer when users want assistance without explicit requests, yet designs rely on untested assumptions about these preferences. We provide behavioral evidence that compositional effort, not urgency, drives AI drafting preferences, shifting design priorities from temporal signals to cognitive load.

We also reveal a perception-behavior gap with measurable consequences. Users rank urgency as most important in self-reports despite its minimal influence, and systems based on stated preferences underperform those that ignore user beliefs. This finding challenges the common practice of soliciting explicit user feedback for system optimization.

Our contributions advance human-centered NLG by: (1) providing behavioral grounding for proactive assistance systems through factorial vignette methodology, (2) demonstrating limits of user introspection where stated preferences mislead design, and (3) showing that interpretable rules derived from revealed preferences match complex ML models and that stated-preference designs underperform naive baselines. As AI becomes more proactive, effective design should prioritize behavioral evidence over intuition, with implications for email assistants, document generators, and conversational tools that infer needs from context.
\newpage
\section{Limitations}
\label{sec:limitations}

\textbf{Controlled experimental setting and scenario coverage.} Our factorial vignette design enables causal identification but abstracts from real-world email complexity. We binarized four dimensions i.e., urgency, effort, sender importance, email type, rather than capturing continuous variation or additional factors like emotional valence or political sensitivity. The approximately 60\% ML accuracy ceiling reflects these constraints. Production systems with richer features such as email content, user history, continuous measures would likely achieve higher performance while maintaining the advantage of behavioral over stated preference weighting.

\textbf{Sample characteristics.} Participants were 50 knowledge workers making frequent email decisions, appropriate for the target population of proactive drafting assistants. We collected data on role distribution (individual contributors, managers), AI usage frequency, AI comfort levels, and email volume to assess generalizability across user segments. However, generalization to populations with systematically different email practices (e.g., customer service roles with heavily templated responses, executives whose correspondence is managed by assistants, non-English communication contexts) requires empirical validation. The perception-behavior gap appears robust across usage frequency and email volume within our sample, but cross-cultural and organizational variation warrants investigation.

\textbf{Task generalizability.} Email drafting involves visible compositional effort (writing, editing, structuring arguments), making effort-to-compose a naturally salient dimension. For different NLG tasks (e.g., summarization requests or query reformulation), effort may manifest differently (comprehension load vs. generation load). However, the general principle that users misidentify decision drivers likely extends beyond email to other proactive assistance contexts.

\textbf{Preference stability.} We measured preferences at a single timepoint. Longitudinal studies would clarify whether preferences evolve with AI adoption, whether early versus experienced users show different factor weightings, and whether revealing the perception gap through usage feedback alters subsequent preferences.

\bibliography{ref}

\newpage
\appendix
\section{Appendix}

\subsection{Email Scenario Descriptions and Generation Prompt}
\label{subsec:scenarios}

\subsubsection{Scenario generation prompt}

\textbf{System Prompt:}

You are a helpful assistant that generates realistic work scenarios. Create natural, contextual scenarios that subtly embed the required dimensions without explicitly labeling them. Participants should be able to infer urgency, effort level, sender importance, and routine/novel nature from context clues rather than direct statements.

Important: Write all scenarios in narrative form without using quotation marks or direct dialogue. Naturally mention how each sender relates to the user's project management role to establish their importance level. Avoid explicit phrases like "this is urgent", "high effort", "routine task", or "novel situation". Let the context convey these qualities. Focus on realistic workplace situations that feel authentic and allow participants to make their own judgments about draft helpfulness.

\textbf{User Prompt:}

Role-Play Profile: Project Manager at a Big Tech Company

You are a project manager at a large technology company, responsible for driving cross-functional initiatives that impact multiple product teams. You have been in this role for three years and currently oversee several projects at different stages of execution. Your work involves coordinating with engineering, design, legal, and communications teams to ensure timely delivery of features and organizational changes.

Your responsibilities include creating and maintaining detailed project plans, managing risks and dependencies, and preparing executive updates. You frequently draft communications for stakeholders, such as status reports, decision summaries, and change announcements. Approvals often require collaboration with legal and compliance teams, and you must ensure all communications adhere to company policies. You also facilitate workshops, track progress against milestones, and escalate issues when necessary.

Communication is a critical part of your role. For executives, you provide concise summaries that highlight decisions, risks, and next steps. For operational teams, you share structured updates with clear timelines and ownership details. You primarily use email and Teams for communication, and you are expected to respond promptly to urgent requests while balancing long-term deliverables.

Task: Generate scenarios for all 16 unique combinations of these dimensions: Urgency (Urgent / Not Urgent), Effort to Compose (High Effort / Low Effort), Sender Importance (Important Sender / Less Critical Sender), Request Type (Routine / Novel).

Guidelines: (1) Convey urgency naturally through context like "by end of day", "ASAP", "needed for tomorrow's meeting" versus "when you get a chance", "no rush", "sometime this week". (2) Show effort level through task complexity rather than stating it explicitly. Lower complexity includes quick confirmations and brief status updates; higher complexity involves multi-team coordination and comprehensive analysis. (3) Establish sender importance through relationship clarity by always specifying how the sender relates to your role. Direct impact senders include your manager, executives over your projects, and key stakeholders; indirect impact senders include peer managers from other teams and administrative staff. (4) Indicate whether requests are typical or unusual through context like "weekly report" versus "unexpected situation" or "new initiative". (5) Write naturally without explicit dimension labels. (6) Make scenarios realistic and specific with authentic context that would occur in a tech company. (7) Ensure each scenario requires a meaningful response. (8) Write scenarios without quotation marks in narrative form.

After generating all 16 scenarios, the authors independently reviewed each scenario to verify that it accurately reflected the intended dimension levels (urgency, effort, sender importance, and email type). The authors agreed that the generated scenarios matched the factorial design specifications.

\subsubsection{Complete scenario list}

\begin{enumerate}
\item \textbf{Urgent + High Effort + Important Sender + Routine}: Your manager reaches out asking for an updated end-to-quarter project health report that consolidates metrics from every active workstream. They need it finalized by end of day for a leadership review later tonight, and the template is the same one your team uses every quarter.

\item \textbf{Urgent + High Effort + Important Sender + Novel}: The VP of Engineering, who oversees the team building your core product, informs you that the executive board wants an impact analysis immediately on how yesterday's regulatory announcement will affect your current launch plan. They request a detailed scenario assessment and mitigation options by tomorrow morning.

\item \textbf{Urgent + High Effort + Less Critical Sender + Routine}: A peer project manager from the HR technology team asks you this morning to produce a comprehensive dependency summary for shared resources. They explain it's blocking their integration test kickoff scheduled for later today.

\item \textbf{Urgent + High Effort + Less Critical Sender + Novel}: The facilities operations coordinator contacts you saying a major system outage has affected some of the collaboration tools used by your project teams. They ask you to help compile an immediate contingency plan with alternative workflows to share with impacted groups before the next working session.

\item \textbf{Urgent + Low Effort + Important Sender + Routine}: Your director pings you requesting the latest timeline slide for an executive call that starts in an hour. They mention the earlier version is fine, but just needs the updated completion percentages inserted.

\item \textbf{Urgent + Low Effort + Important Sender + Novel}: A senior product lead messages you asking for a quick confirmation on whether your team can accommodate an unplanned customer feedback session tomorrow. They need a response now so they can commit to the client during an ongoing call.

\item \textbf{Urgent + Low Effort + Less Critical Sender + Routine}: An analyst from another department emails asking you to resend yesterday's stakeholder deck because they can't access it in the shared folder. They need it now to join a call that starts in a few minutes.

\item \textbf{Urgent + Low Effort + Less Critical Sender + Novel}: A facilities admin contacts you during a building-wide power fluctuation, wanting to know if any hardware dependencies in your programs need immediate communication to engineering. They ask for a quick yes or no so they can finalize their outage alert message.

\item \textbf{Not Urgent + High Effort + Important Sender + Routine}: Your director asks you to prepare the full quarterly portfolio summary for their upcoming strategy review next week. It's the same in-depth report you've created in past quarters, combining KPIs, risk logs, and milestone charts.

\item \textbf{Not Urgent + High Effort + Important Sender + Novel}: The Chief Legal Officer requests that you draft a comprehensive transition plan outlining how upcoming compliance rules will be integrated into all your product streams. They emphasize that this initiative is new to the organization and due end of next month, so you have time to propose a structured approach.

\item \textbf{Not Urgent + High Effort + Less Critical Sender + Routine}: A peer PM from an internal tools project asks if you can produce a consolidated view of shared resource usage between your project and theirs. They note it isn't blocking any current work and suggest having something ready in a few weeks.

\item \textbf{Not Urgent + High Effort + Less Critical Sender + Novel}: The communications specialist for the employee engagement program reaches out for help designing a cross-team workshop to brainstorm ways of showcasing successful product launches internally. They mention it's a new initiative and would appreciate speaking with you later this month.

\item \textbf{Not Urgent + Low Effort + Important Sender + Routine}: Your manager emails asking you to confirm whether last sprint's milestone progress is already reflected in the dashboard. They say there's no rush and to let them know anytime this week.

\item \textbf{Not Urgent + Low Effort + Important Sender + Novel}: A senior engineering director asks if you can provide a quick paragraph describing your project's long-term architecture considerations for a vision document they're drafting. They mention having it next week is fine.

\item \textbf{Not Urgent + Low Effort + Less Critical Sender + Routine}: An administrative coordinator pings you for the standard monthly headcount figure for your project team. They clarify that they're compiling next month's seating plan and you can update them whenever convenient this week.

\item \textbf{Not Urgent + Low Effort + Less Critical Sender + Novel}: A colleague working on an internal culture committee explains they're collecting brief success stories from various programs for a new company blog. They ask for a short summary of a recent milestone your project achieved whenever you have a moment in the coming days.
\end{enumerate}

\subsection{Thematic Analysis Methodology and Secondary Findings}
\label{subsec:thematic_methods}

We conducted exploratory thematic analysis to complement the quantitative preference modeling. This analysis examined textual responses for two research questions. The first research question investigated what email attributes drive users' preferences for AI drafting assistance, analyzing 750 in-the-moment justifications provided during pairwise comparisons. The second research question examined whether users accurately perceive what drives their preferences, analyzing 50 post-task survey reflections where participants retrospectively described their decision-making processes.

\subsubsection{Five-phase LLM-assisted coding process}

We employed a systematic five-phase approach adapted from traditional qualitative methods but operationalized through LLM assistance to enable broader coverage while maintaining analytical rigor.

\textbf{Phase 1: Initial coding.} The LLM analyzed a representative sample using inductive coding with open-ended prompts designed to let themes emerge naturally. For the first research question on preference drivers, we sampled 150 of 750 justifications to balance coverage and manageability, representing 20\% of all pairwise comparisons. For the second research question on metacognitive accuracy, we analyzed all 50 survey responses given the smaller dataset. The prompt explicitly instructed the model to let themes emerge from the responses themselves without forcing responses into predefined categories. If the model noticed themes that did not fit common patterns, it was instructed to include them as distinct codes. Temperature was set to 0.3 for consistent coding while allowing theme discovery.

\textbf{Phase 2: Codebook development.} Initial themes were consolidated into a structured codebook with hierarchical organization, clear definitions, inclusion criteria, exclusion criteria, and example quotes for each code. This phase used temperature 0.3 to balance systematicity with synthesis.

\textbf{Phase 3: Systematic coding.} The codebook was applied to all responses using very low temperature of 0.1 to ensure maximum consistency. Responses for the first research question were processed in batches of 50 to accommodate the richer pairwise comparison context where each response included two email scenarios, the preference choice, and the justification text. All 50 responses for the second research question were coded in one batch given the simpler data structure with only open-ended text.

\textbf{Phase 4: Parsing and structuring.} Coded results were parsed, merged with original responses, and analyzed for theme frequencies, co-occurrence patterns, and distribution statistics.

\textbf{Phase 5: Synthesizing findings.} The authors synthesized findings using LLM-generated summaries of theme distributions, sample quotes, and codebook patterns. For the second research question specifically, prompts emphasized perception-behavior gap analysis by asking about factors users over-report, under-report, or misattribute when describing their decision-making. The LLM provided organized summaries at temperature 0.4 for moderate analytical creativity while staying grounded in data, which the authors then interpreted to identify key patterns and implications.

All phases used a fixed random seed of 42 for reproducibility. The authors reviewed and confirmed the coding process and outputs at each phase to ensure that results and insights were valid and aligned with the data. Complete codebooks with hierarchical theme structures, coded datasets with theme assignments, and detailed frequency distributions are presented in this appendix (see Appendix~\ref{subsec:thematic_findings} below).

\subsubsection{Complete Thematic Findings}
\label{subsec:thematic_findings}

\textbf{RQ1: In-the-moment justifications during pairwise comparisons}

We analyzed 750 justifications provided immediately after participants made their preference choices during the pairwise comparison task. We identified 12 distinct themes organized into four main categories following the hierarchical codebook structure developed in Phase 2. Total justifications coded: 750. Total theme assignments: 769. Average themes per justification: 1.03. Median: 1. Note: Percentages below represent the proportion of responses mentioning each theme. They sum to 102.5\% (rather than 100\%) because some responses mention multiple themes simultaneously.

\textbf{Complete theme inventory with hierarchical structure}

\textbf{Main Theme 1: Time and efficiency drivers (31.5\% combined)}
\begin{itemize}
\item \textbf{Urgency and time pressure} (20.3\%) cited tight deadlines or urgent turnaround as reason for AI use
\item \textbf{Desire to save time} (11.2\%) emphasized efficiency or reducing effort without explicit urgency
\end{itemize}

\textbf{Main Theme 2: Task characteristics (40.7\% combined)}
\begin{itemize}
\item \textbf{Complexity of drafting} (14.7\%) indicated task is complex, multi-step, or requires synthesis
\item \textbf{Task simplicity – doesn't need AI} (12.8\%) determined task is too simple for AI to add value
\item \textbf{Template or historical context} (10.1\%) noted existing format, template, or prior example that AI can leverage
\item \textbf{Rote or tedious work} (3.1\%) identified repetitive, low-value, or boring tasks
\end{itemize}

\textbf{Main Theme 3: Risk and trust considerations (6.1\% combined)}
\begin{itemize}
\item \textbf{Avoiding high-risk or sensitive tasks} (4.8\%) avoided AI for sensitive, high-stakes, or regulatory content
\item \textbf{Ease of verification} (1.3\%) indicated ability to easily check correctness, reducing risk
\end{itemize}

\textbf{Main Theme 4: Perceptions of AI role and capability (24.3\% combined)}
\begin{itemize}
\item \textbf{Perceived AI capability} (14.3\%) believed AI can perform well due to clear requirements or structured tasks
\item \textbf{Alignment with AI strengths} (4.5\%) identified task matches AI's known strengths like summarization or structuring
\item \textbf{Brainstorming or creative help} (3.9\%) saw AI as helpful for generating ideas or creative input
\item \textbf{Personal writing preference} (1.6\%) chose AI due to personal discomfort or dislike of writing tasks
\end{itemize}

\textbf{Co-occurrence patterns reveal contextual trust.} Urgency and time pressure frequently co-occurred with complexity of drafting, indicating users conceptualize high-stakes scenarios as combining both time pressure and cognitive load. Notably, perceived AI capability clustered with template or historical context, suggesting users calibrate their trust in AI based on task predictability rather than holding uniform beliefs about AI capability. When prior examples exist, users explicitly justify AI use with phrases like ``AI seems like it would be able to accurately write this'' and ``format already existing so human validation will be faster.'' This conditional trust pattern suggests personalized AI recommendations should emphasize template availability as a trust signal.

\textbf{Verification as risk mitigation strategy.} Ease of verification appeared alongside avoiding high-risk tasks, revealing a sophisticated mental model where users do not simply avoid AI for high-stakes tasks but rather assess whether verification is feasible. Example quotes include ``It's easy for me to verify correctness'' and ``This is a place where AI can quickly get things in the right format and it's easy for me to verify.'' This suggests that risk-averse users may adopt AI if verification workflows are streamlined, reframing the design challenge from making AI more accurate to making verification effortless.

\textbf{Template and historical context as dual function.} Template or historical context (10.1\%) serves two distinct functions in justifications: reducing perceived effort (e.g., ``uses the same report I've used before, and can save me time'') and increasing AI trust through prior examples and predictable formats. This dual function makes templates particularly powerful triggers for AI adoption, as they simultaneously lower perceived risk and increase perceived benefit.

\textbf{Personal writing discomfort as niche factor.} Personal writing preference (1.6\%) emerged as a distinct but infrequent theme, with users mentioning concerns like ``My style of writing is a little less formal'' and tone calibration. While rare, this suggests a user segment that seeks AI not for efficiency but for stylistic support. This niche use case differs from the dominant effort-reduction pattern and may benefit from different feature prioritization (tone suggestions vs. quick drafts).

\textbf{Secondary design implications.} Beyond the primary implications discussed in the main paper, these secondary findings suggest: (1) Template-based triggers should be emphasized in UI as both trust and efficiency signals. (2) Verification dashboards showing confidence scores and highlighting AI-generated content could expand adoption among risk-averse users. (3) Cognitive load detection (typing pauses, revision patterns) may be more predictive than self-reported urgency. (4) Segmented features for niche users (tone adjustment, brainstorming mode) complement core efficiency features.

\textbf{RQ2: Post-task survey reflections on decision-making}

We analyzed 50 survey responses to the open-ended question asking what factors participants considered when deciding whether an email is worth drafting. We identified 13 distinct themes (excluding "None" responses). Total responses coded: 50. Average themes per response: 2.82. Median: 3. Note: Percentages below represent the proportion of responses mentioning each theme. They sum to more than 100\% (282\%) because each response typically mentions multiple themes.

\textbf{Complete theme inventory}
\begin{itemize}
\item \textbf{Urgency / time sensitivity} (66\%) referenced deadlines, time-critical nature, or need for quick turnaround
\item \textbf{Effort vs. efficiency} (32\%) considered effort required to draft manually versus using AI, including time savings
\item \textbf{Data availability and inputs} (28\%) assessed whether necessary information or data is readily available
\item \textbf{Complexity of content} (26\%) evaluated how complicated or technical the subject matter is
\item \textbf{Audience / stakeholder importance} (24\%) considered who the email is for, including seniority and role
\item \textbf{Task type or nature} (22\%) distinguished inherent nature of request like informational versus action-oriented
\item \textbf{Fit for AI / AI capability} (20\%) assessed whether task aligns with AI's strengths or AI can handle complexity
\item \textbf{Importance / impact of email} (20\%) evaluated stakes, visibility, or potential consequences
\item \textbf{Length and detail of email} (18\%) considered expected size and amount of content required
\item \textbf{Accuracy and quality requirements} (12\%) weighed need for correctness, reliability, and professional tone
\item \textbf{Need for personalization or nuance} (8\%) assessed need for personal touch or emotional tone
\item \textbf{Prior experience / familiarity} (6\%) drew on past experience with similar emails or AI performance
\end{itemize}

\textbf{Retrospective elaboration as rationalization mechanism.} The average themes per response increased from 1.03 in immediate justifications to 2.82 in retrospective accounts. This pattern may reflect reconstructive memory rather than merely increased thoroughness, where users generate more elaborate multi-factorial explanations when asked to reflect, consistent with confabulation literature. The consistency in top themes across users (urgency 66\%, effort 32\%) despite individual variation in actual behavior suggests shared cultural scripts about "good decision-making" rather than genuine introspection.

\textbf{Data availability as gating factor.} Data availability and inputs emerged strongly in retrospective accounts (28\%) but was absent from the original four-dimension experimental design. This theme reveals users recognize information completeness as a prerequisite, with quotes like "how much chasing people I'd need to do to get the data" and "Amount of data that needs to be consolidated." This suggests a two-stage decision process: first assessing feasibility (do I have the data?), then assessing desirability (is AI help worth it?). Production systems should surface data completeness signals proactively.

\textbf{Limited metacognitive learning about AI preferences.} Despite completing 15 pairwise comparisons with AI drafting scenarios, participants rarely referenced prior experience or familiarity in their reflections (6\%). Users seldom mentioned learning from the task itself or updating beliefs about AI capability. This suggests that metacognitive learning about AI preferences develops slowly, and systems may need explicit feedback loops showing users their revealed patterns to accelerate this calibration.

\textbf{Quality concerns as professional identity.} Accuracy and quality requirements (12\%) and need for personalization or nuance (8\%) emerged in retrospective reflections despite not being experimentally manipulated dimensions. These themes reflect professional identity maintenance, with users emphasizing "reliability of accuracy," "quality standards," and "personal touch needed." Their presence in retrospective accounts but absence as behavioral predictors suggests they serve as post-hoc justifications for professionalism rather than actual decision drivers.

\textbf{Task type taxonomy as mental model.} Task type or nature (22\%) reveals users mentally categorize emails along multiple dimensions beyond the experimental variables: "informational vs. action-oriented," "repackaging existing content," "request was for a text response or me just taking action." This richer taxonomy suggests the four-factor experimental model captures primary drivers but users perceive additional nuances that may matter for edge cases or specific workflows.

\textbf{Secondary design implications.} These secondary findings suggest: (1) Data completeness indicators should be visible before drafting ("You have all needed information" or "Missing: Q3 budget figures"). (2) Usage pattern dashboards could accelerate metacognitive calibration ("Your AI requests: 78\% high-effort emails, 12\% urgent emails"). (3) Quality assurance features (tone matching, fact-checking) address post-hoc professional concerns even if not behavioral drivers. (4) Task type detection could route emails to specialized AI capabilities (informational summary vs. action proposal generation).

\subsection{ML Model Implementation Details}
\label{subsec:ml_details}

This section provides complete implementation details for the ML models used in \S\ref{sec:rq3}.

\subsubsection{Data preparation}

\textbf{Feature construction.} Each of the 750 pairwise comparisons was converted into a binary classification instance. For each comparison between scenarios $i$ and $j$, we created features based on the difference in factor levels: $\Delta$Urgency, $\Delta$Effort, $\Delta$Sender, $\Delta$Type $\in$ \{-1, 0, +1\}. We also included all six two-way interaction terms: $\Delta$Urgency $\times$ $\Delta$Effort, $\Delta$Urgency $\times$ $\Delta$Sender, $\Delta$Urgency $\times$ $\Delta$Type, $\Delta$Effort $\times$ $\Delta$Sender, $\Delta$Effort $\times$ $\Delta$Type, and $\Delta$Sender $\times$ $\Delta$Type. The target variable was binary: 1 if scenario $i$ was preferred, 0 if scenario $j$ was preferred. This resulted in 10 features per instance.

\textbf{Preprocessing.} All features were standardized using StandardScaler (zero mean, unit variance) to ensure comparable scales across models. No dimensionality reduction was applied given the already compact feature set.

\textbf{Train-test split.} We used stratified 5-fold cross-validation to ensure balanced class distribution across folds and robust performance estimates. Random seed was set to 42 for reproducibility.

\subsubsection{Model architectures and hyperparameters}

\textbf{Logistic Regression.} Standard L2-regularized logistic regression with regularization strength $C = 1.0$. Solver: lbfgs. Maximum iterations: 1000. Class weights: balanced to account for slight class imbalance (55.2\% prefer assistance).

\textbf{Random Forest.} Ensemble of decision trees with the following hyperparameters tuned via grid search: number of estimators = 100, max depth = 3, min samples split = 20, min samples leaf = 10. These shallow tree constraints were chosen to prevent overfitting given modest dataset size (750 instances). Feature importance was extracted using Gini importance averaged across all trees.

\textbf{Gradient Boosting.} Gradient-boosted decision trees with: learning rate = 0.1, number of estimators = 100, max depth = 3, min samples split = 20, min samples leaf = 10, subsample = 0.8. Similar depth constraints as Random Forest for overfitting prevention.

\textbf{Neural Network.} Shallow feedforward architecture to prevent overfitting:
\begin{itemize}[itemsep=0pt, parsep=0pt, topsep=0pt]
\item Input layer: 10 features (4 main effects + 6 interactions)
\item Hidden layer 1: 16 units, ReLU activation
\item Hidden layer 2: 8 units, ReLU activation  
\item Output layer: 1 unit, sigmoid activation
\end{itemize}

Training configuration:
\begin{itemize}[itemsep=0pt, parsep=0pt, topsep=0pt]
\item Optimizer: Adam with learning rate = 0.001
\item Loss function: Binary cross-entropy
\item Batch size: 32
\item Epochs: 50 with early stopping (patience = 10, monitoring validation loss)
\item Dropout: 0.3 after each hidden layer for regularization
\item L2 regularization: weight decay = 0.01
\end{itemize}

\subsubsection{Rationale for shallow architectures}

With only 750 training instances and 10 features, we prioritized generalization over model complexity. Initial experiments with deeper networks (3-4 hidden layers with 32-64 units) showed severe overfitting: training accuracy reached 75-80\% while validation accuracy plateaued at 58-60\%. Shallow architectures (2 hidden layers with 16 and 8 units) provided better bias-variance tradeoff, achieving 61-62\% validation accuracy with minimal train-validation gap (3-4 percentage points). Similarly, tree-based models with max depth $>$ 5 showed overfitting, justifying the depth = 3 constraint.

This design choice aligns with the paper's core finding: simple behavioral rules (61.3\% accuracy) match complex ML models (60.8\% mean accuracy), suggesting the predictive ceiling reflects data constraints rather than insufficient model capacity.

\subsubsection{Performance evaluation}

\begin{table}[h]
\centering
\small
\begin{tabular}{l@{\hskip 0.2cm}c@{\hskip 0.2cm}c@{\hskip 0.2cm}c@{\hskip 0.2cm}c@{\hskip 0.2cm}c@{\hskip 0.2cm}c@{\hskip 0.2cm}c@{\hskip 0.2cm}c}
\toprule
& \multicolumn{4}{c}{\textbf{Full Features (10)}} & \multicolumn{4}{c}{\textbf{Main Effects (4)}} \\
\cmidrule(lr){2-5} \cmidrule(lr){6-9}
Model & Acc. & Prec. & Rec. & F1 & Acc. & Prec. & Rec. & F1 \\
\midrule
Logistic Regression & 0.587 & 0.627 & 0.691 & 0.657 & 0.580 & 0.624 & 0.664 & 0.643 \\
Random Forest & 0.608 & 0.607 & 0.882 & 0.719 & 0.605 & 0.614 & 0.855 & 0.714 \\
Gradient Boosting & 0.588 & 0.617 & 0.833 & 0.709 & 0.600 & 0.615 & 0.848 & 0.713 \\
Neural Network & 0.599 & 0.619 & 0.785 & 0.692 & 0.603 & 0.608 & 0.819 & 0.698 \\
\midrule
\textbf{Mean} & \textbf{0.595} & \textbf{0.618} & \textbf{0.798} & \textbf{0.694} & \textbf{0.597} & \textbf{0.615} & \textbf{0.797} & \textbf{0.692} \\
\bottomrule
\end{tabular}
\caption{Complete ML model performance metrics. Main paper (Table~\ref{tab:ml_performance}) shows accuracy and F1 only for brevity.}
\label{tab:ml_performance_full}
\end{table}

\newpage
\onecolumn
\subsection{Study Interface Screenshots and Exit Survey}
\label{subsec:exit_survey}

\begin{figure}[!h]
\centering
\includegraphics[width=0.7\textwidth]{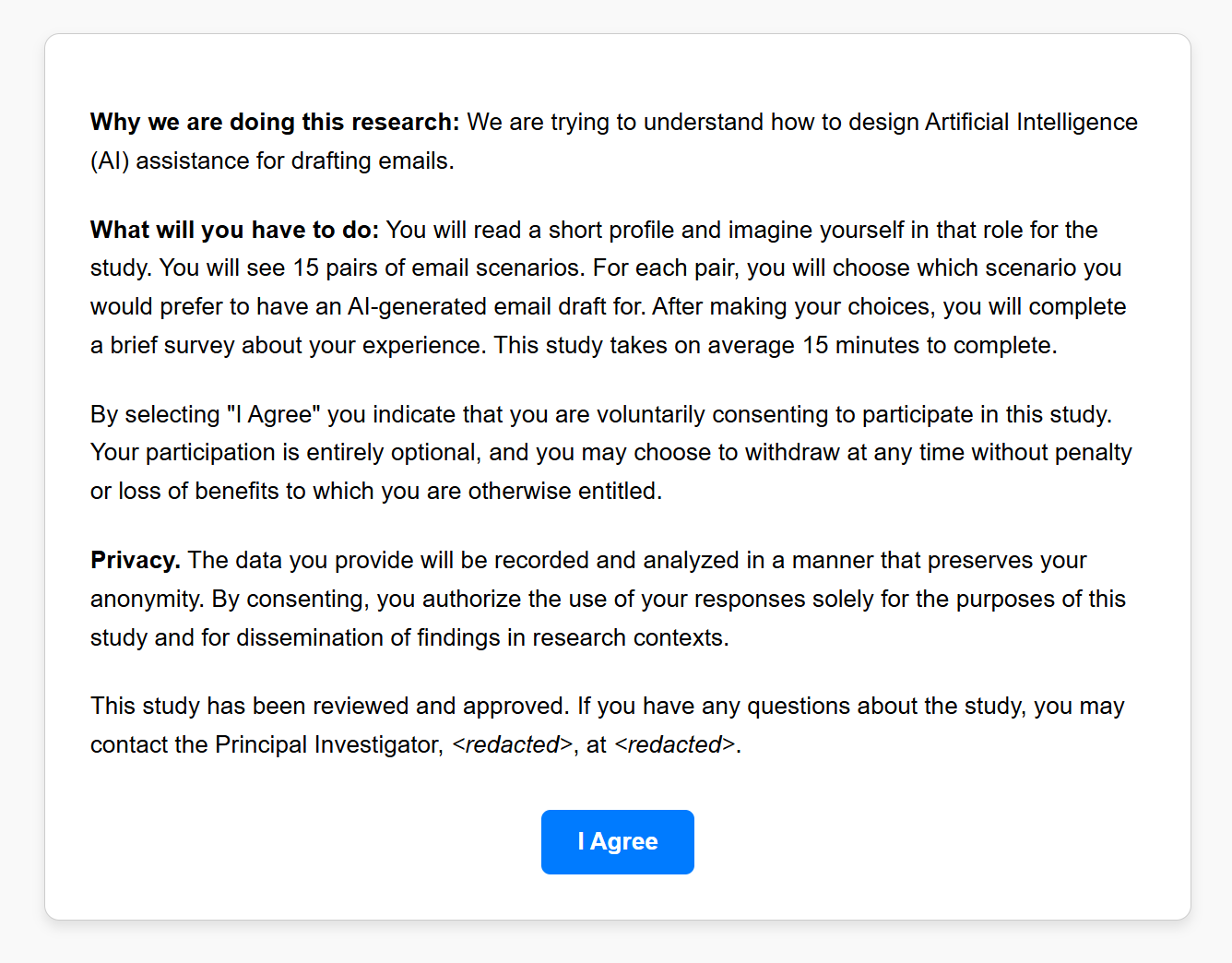}
\caption{Informed consent page presented at the beginning of the study. Participants reviewed study purpose, procedures, risks, benefits, and data handling before providing consent to participate.}
\label{fig:consent}
\end{figure}

\begin{figure}[!h]
\centering
\includegraphics[width=0.7\textwidth]{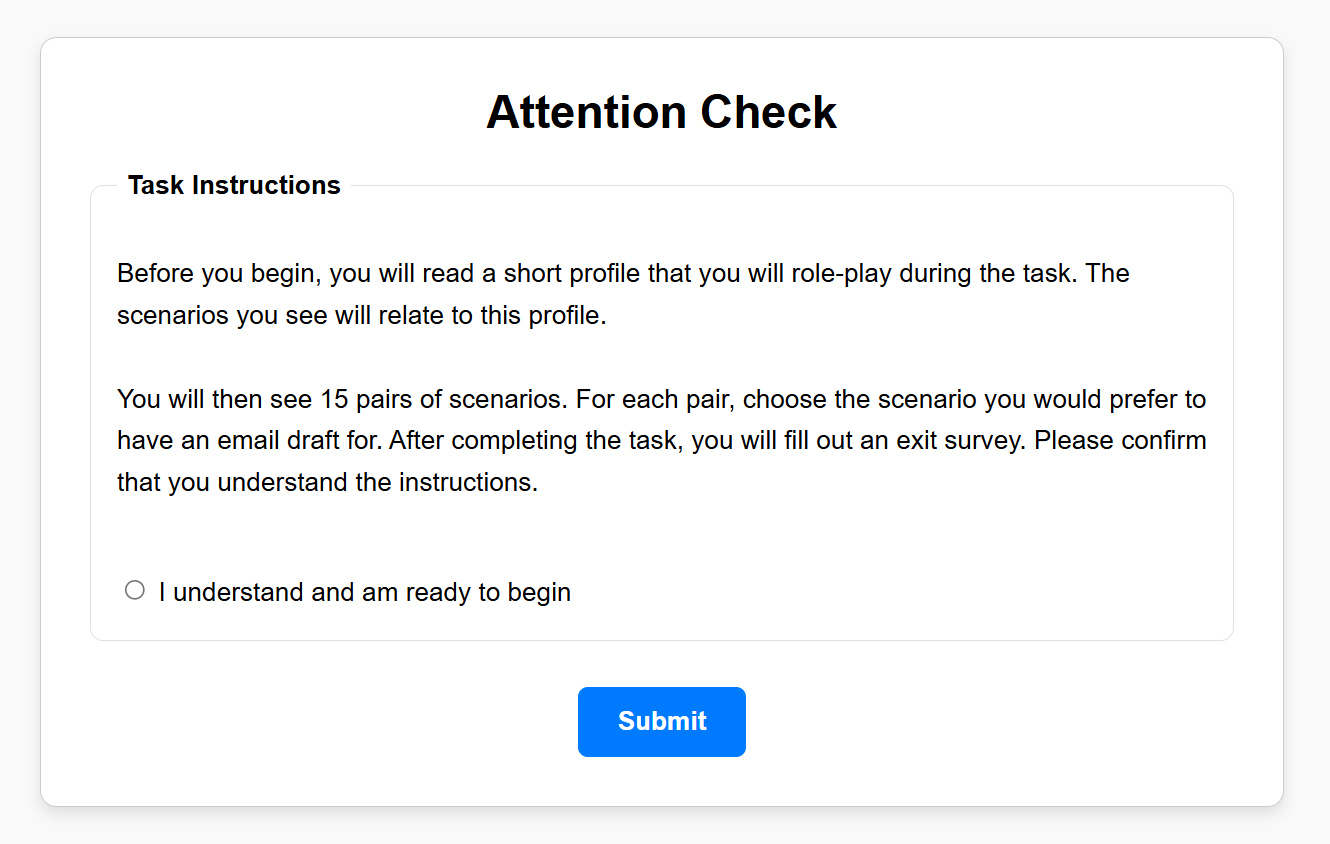}
\caption{Task instructions screen explaining the pairwise comparison task. Participants were informed they would see pairs of email scenarios and select which email they would prefer AI drafting assistance for, then provide justification for their choice.}
\label{fig:task_instructions}
\end{figure}

\begin{figure}[!h]
\centering
\includegraphics[width=0.65\textwidth]{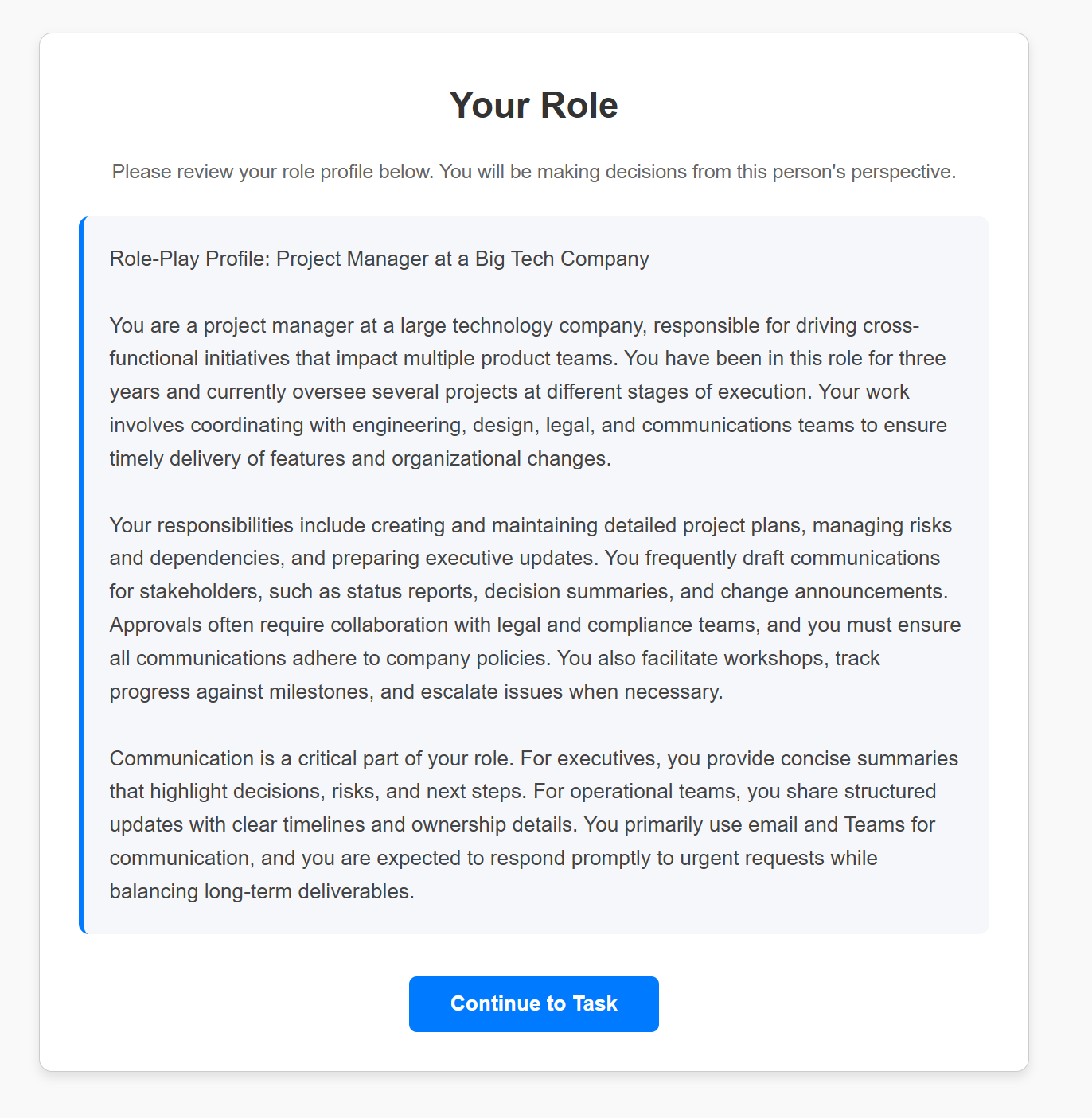}
\caption{Role-play profile assignment. Participants were instructed to adopt the role of a project manager at a large technology company to ground their judgments in a consistent professional context across all scenarios.}
\label{fig:role_play_profile}
\end{figure}

\begin{figure}[!h]
\centering
\includegraphics[width=0.65\textwidth]{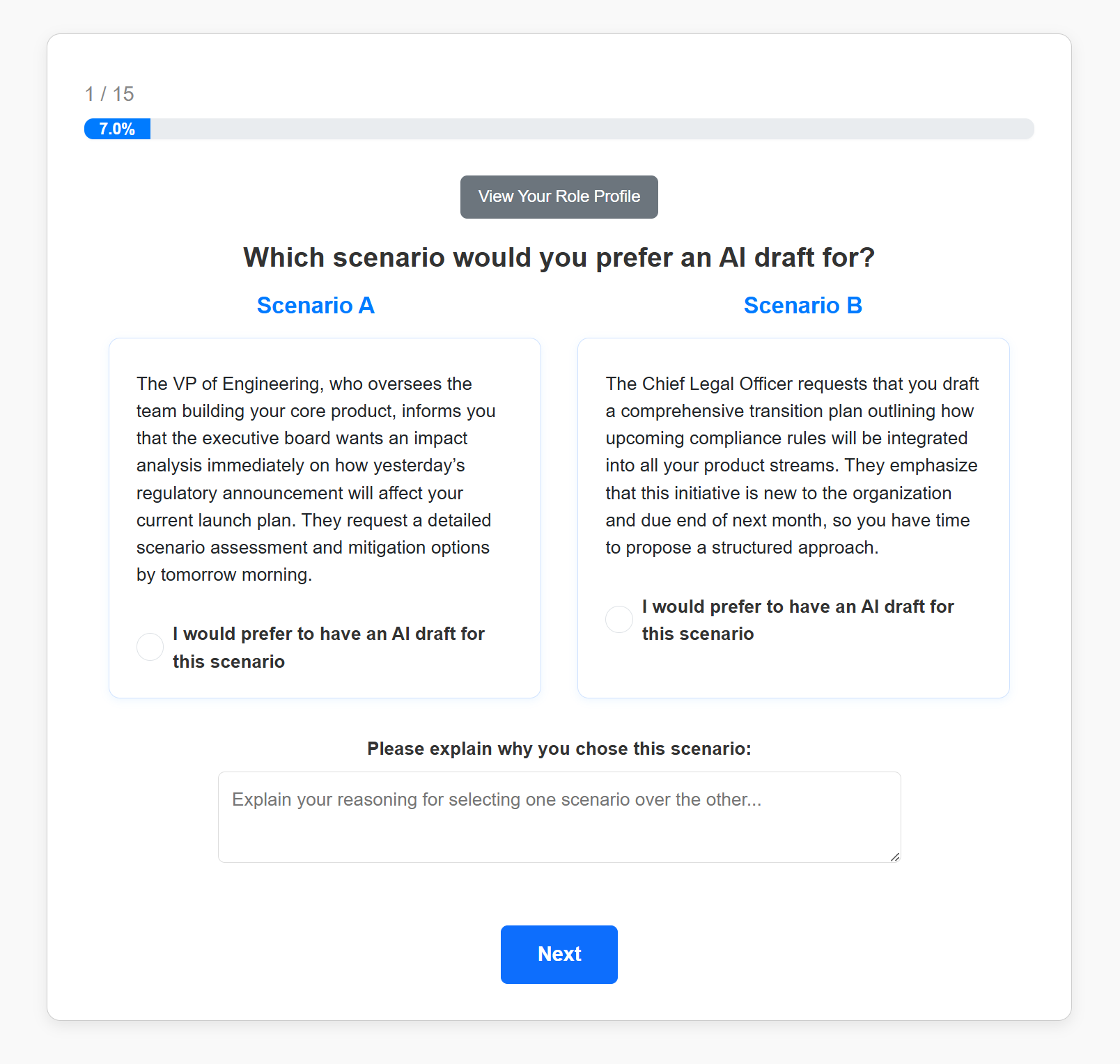}
\caption{Pairwise comparison interface showing the main task. Participants viewed two email scenarios side-by-side, selected which email they would prefer AI drafting assistance for, and provided free-text justification for their choice. Each participant completed 15 such comparisons.}
\label{fig:sbs_comparison}
\end{figure}

\begin{figure}[!h]
\centering
\includegraphics[width=0.7\textwidth]{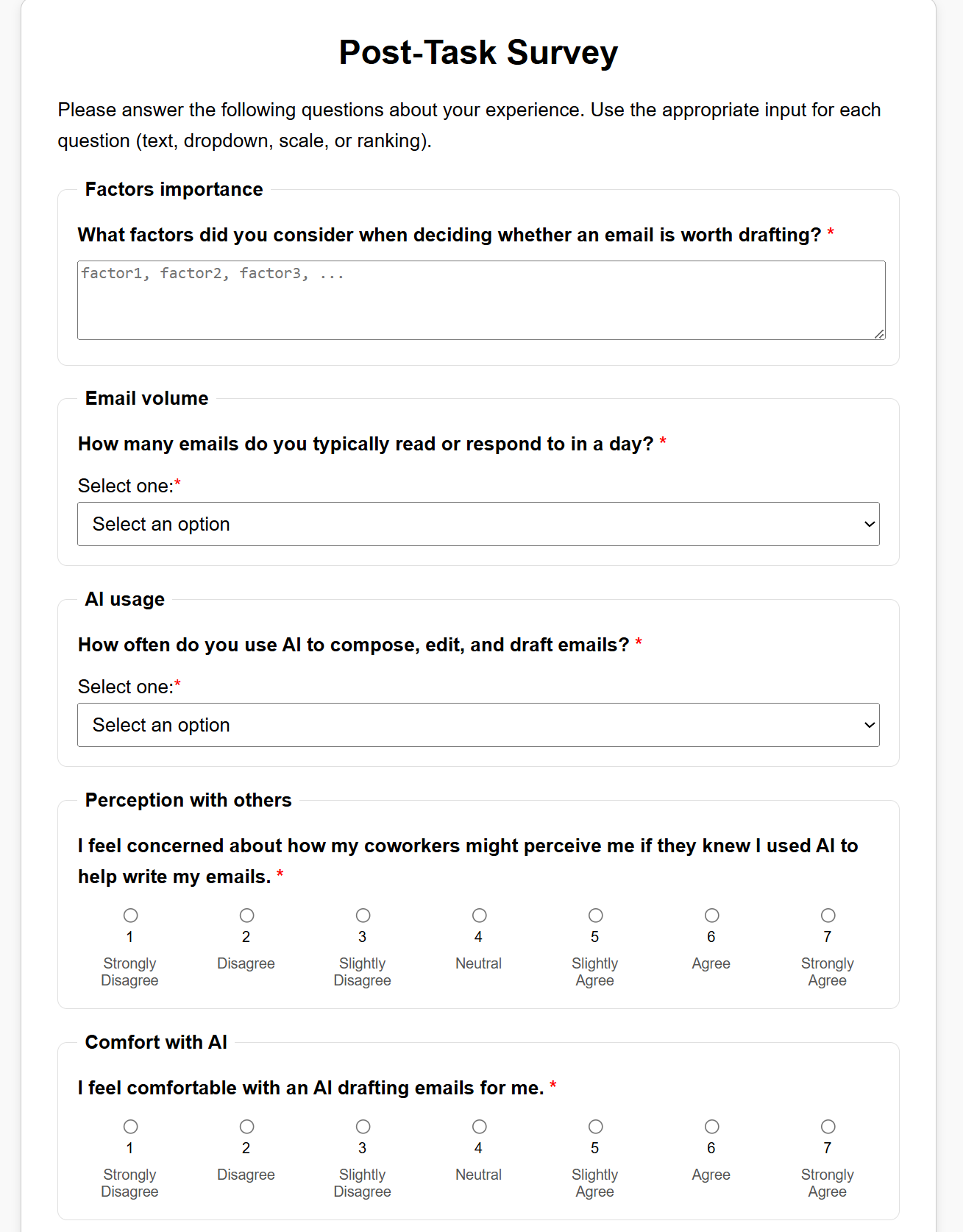}
\caption{Exit survey (screenshot part 1) collecting open-ended reflections on decision factors, email volume, AI usage frequency, and AI attitudes. The open-ended question asked participants to describe what factors they considered when deciding whether an email is worth drafting.}
\label{fig:survey_part1}
\end{figure}

\begin{figure}[!h]
\centering
\includegraphics[width=0.7\textwidth]{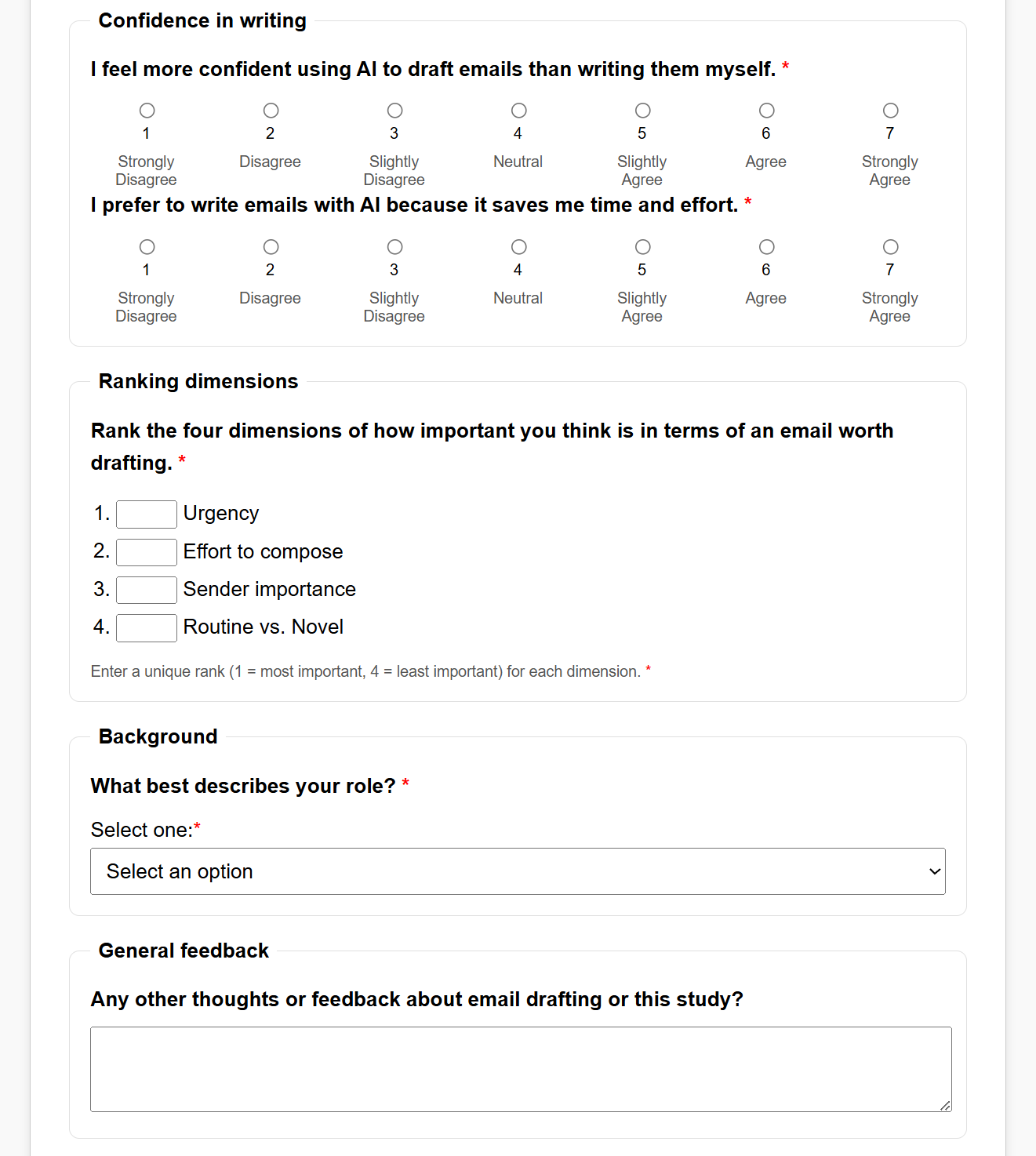}
\caption{Exit survey (screenshot part 2) collecting forced rankings of the four dimensions (Urgency, Effort to Compose, Sender Importance, Email Type) from most to least important in participants' decision-making, along with current role and optional feedback.}
\label{fig:survey_part2}
\end{figure}

\clearpage
\newpage
\subsection{Individual Differences in Preference Patterns}
\label{subsec:appendix_individual_differences}

\begin{figure}[!h]
\centering
\includegraphics[width=0.6\textwidth]{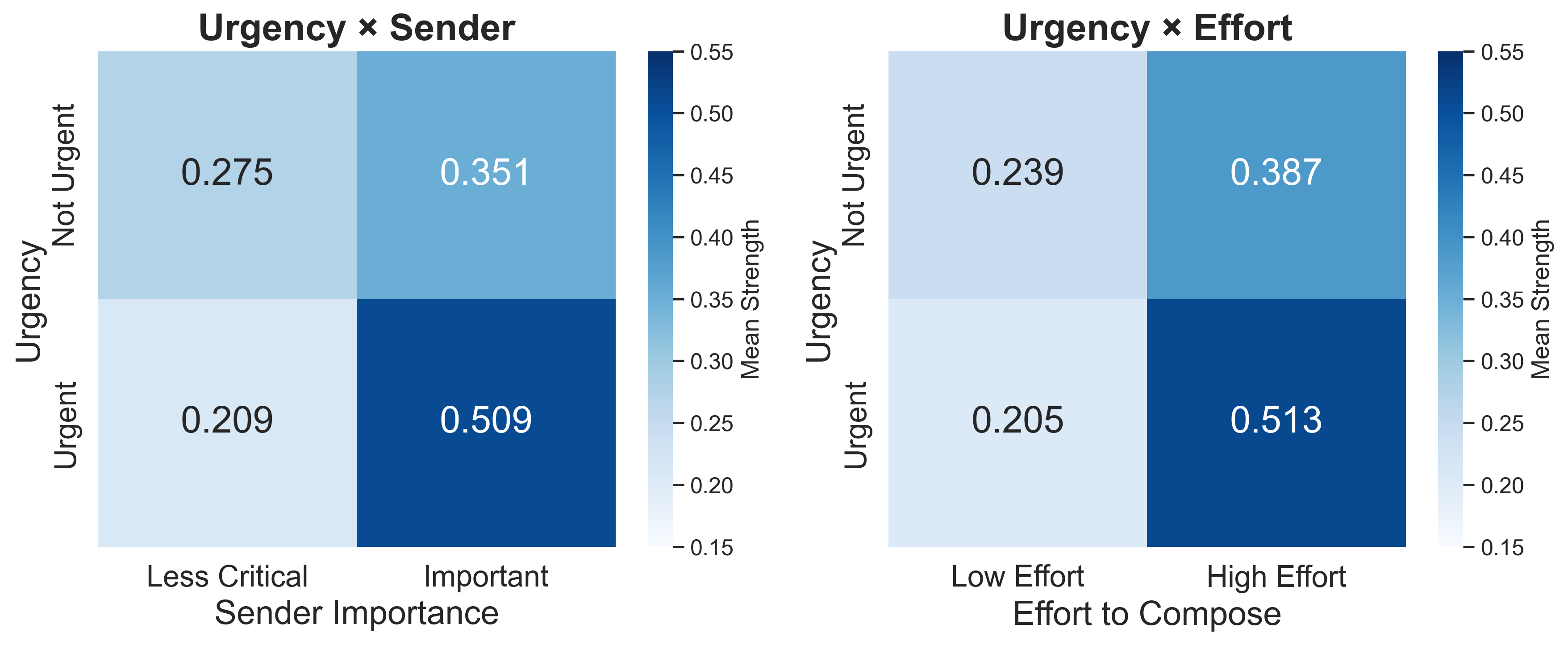}
\caption{Remaining pairwise dimension interactions showing mean Bradley-Terry preference strengths. Urgency $\times$ Sender and Urgency $\times$ Effort show relatively flat patterns, confirming urgency's weak main effect ($\rho \approx 0$) does not emerge through strong interactions with other dimensions.}
\label{fig:dimension_interactions_appendix}
\end{figure}

\begin{figure*}[!h]
\centering
\includegraphics[width=0.8\textwidth]{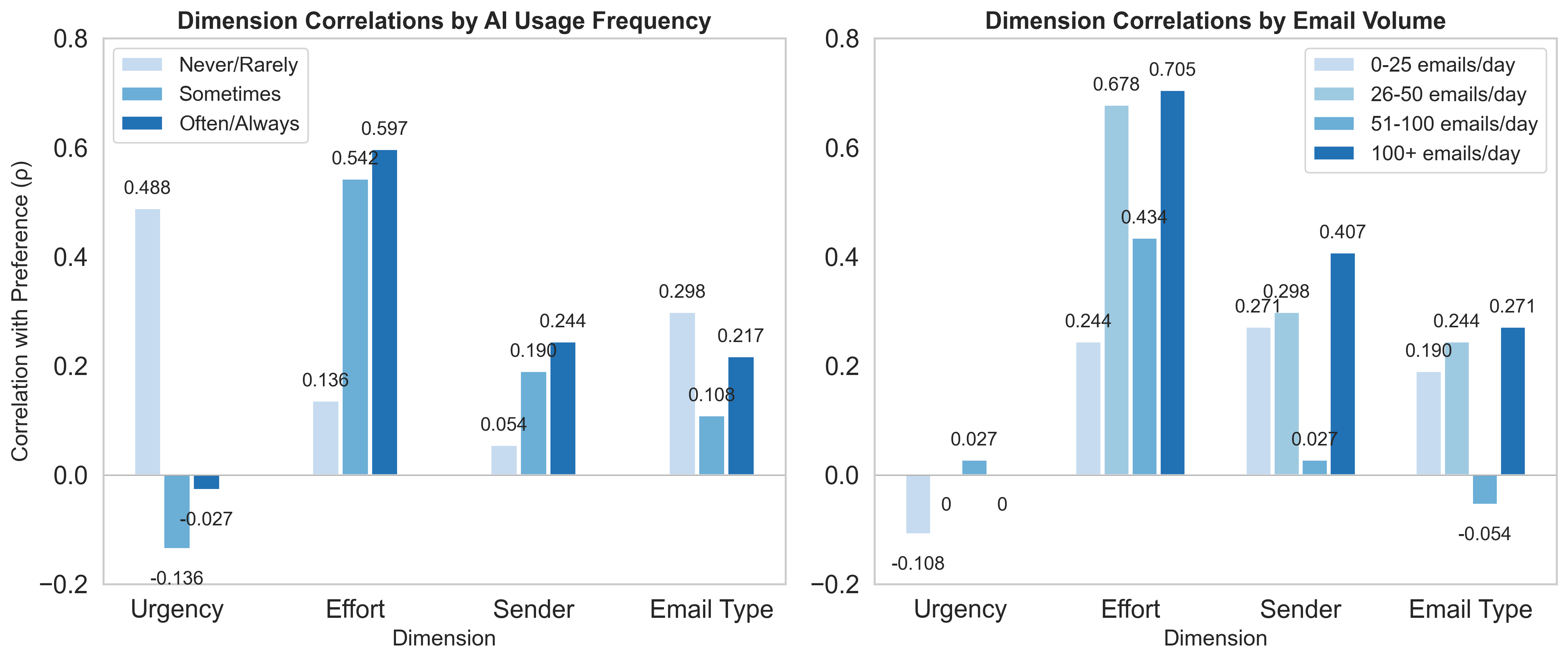}
\caption{Individual variation in preference patterns across user segments. Left panel shows dimension correlations by AI usage frequency; right panel shows correlations by email volume. Effort dominates consistently across all groups (blue bars remain high), while Email Type (routine vs. novel preference) exhibits meaningful variation. Experienced AI users show weaker routine preference ($\rho = 0.217$) compared to infrequent users ($\rho = 0.298$), and high-volume users (100+ emails/day) show stronger routine preference ($\rho = 0.271$) than moderate-volume users. This demonstrates that core preference drivers (effort) generalize broadly while secondary factors (email type familiarity) vary with user experience.}
\label{fig:individual_variation}
\end{figure*}

\begin{figure*}[!h]
\centering
\includegraphics[width=0.9\textwidth]{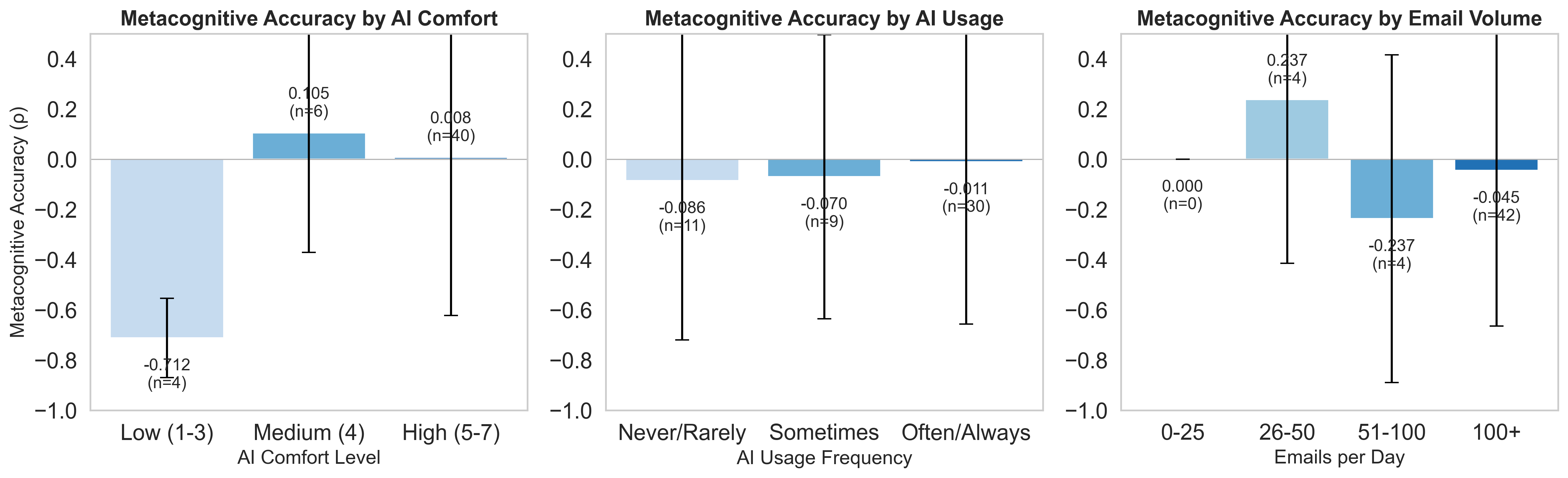}
\caption{Individual differences in metacognitive accuracy (correlation between stated and revealed preferences) across user characteristics. Left panel shows AI comfort level is the key predictor, with high-comfort users showing positive alignment ($\rho = 0.113$) while low-comfort users show inverted understanding ($\rho = -0.550$). Middle and right panels show AI usage frequency and email volume do not significantly predict metacognitive accuracy, indicating psychological readiness matters more than behavioral experience. Error bars represent standard deviation across users in each group.}
\label{fig:metacognitive_accuracy}
\end{figure*}

\end{document}